\documentclass[10pt,twocolumn,letterpaper]{article}

\usepackage[pagenumbers]{cvpr} 

\usepackage{multirow}
\usepackage{makecell}  
\usepackage{colortbl}  
\usepackage{arydshln}  
\newcommand{\dashrule}[1]{\noalign{\vskip 1pt}\cdashline{#1}\noalign{\vskip 2pt}}  

\newcommand{\modelname}{{IDSplat2}}
\newcommand{\ours}{Ours}
\newcommand{\mhr}{MHR}
\newcommand{\sambody}{SAM~3D~Body}
\newcommand{\cmark}{\checkmark}
\newcommand{\xmark}{$\times$}

\newcommand{\parsection}[1]{\noindent\textbf{#1:}}
\usepackage{amssymb}
\usepackage{tabulary}
\definecolor{tabfirst}{rgb}{1, 0.7, 0.7} 
\definecolor{tabsecond}{rgb}{1, 0.85, 0.7} 
\definecolor{tabthird}{rgb}{1, 1, 0.7} 

\usepackage{comment}

\definecolor{cvprblue}{rgb}{0.21,0.49,0.74}
\usepackage[pagebackref,breaklinks,colorlinks,allcolors=cvprblue]{hyperref}

\def\confName{CVPR}
\def\confYear{2027}

\title{Zero-shot Instance-decomposed 3D Gaussian Splatting \\using a Parametric Body Model for Pedestrians}

\author{Mahan Rafidashti$^{1,2}$
\quad Maryam Fatemi$^{1}$
\quad Lars Hammarstrand$^{2}$
\\
\normalsize$^1$Zenseact \hspace{0.8cm} $^2$Chalmers University of Technology \hspace{0.8cm}\\
{\tt\small \{firstname.lastname\}@\{zenseact.com, chalmers.se\}}
}

\begin{document}
\twocolumn[{
    \maketitle
    \vspace{-10mm}
    \begin{center}
        \captionsetup{type=figure}
        \includegraphics[width=0.99\textwidth,trim={0cm 0cm 0cm 0cm},clip]{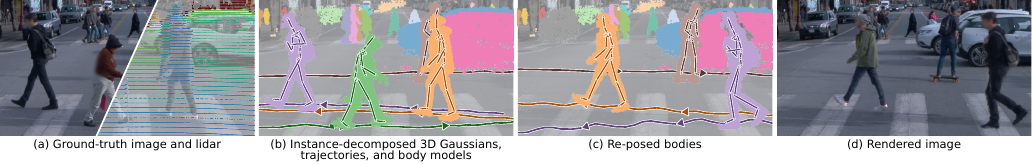}
        \vspace{-2mm}
        \captionof{figure}{Given camera images and lidar point clouds (a), we reconstruct the scene as instance-decomposed 3D Gaussians with pedestrian body models (b), re-pose the pedestrians (c), and render the edited scene (d).}
        \label{fig:topfigure}
    \end{center}
}]
\maketitle
\begin{abstract}
The ability to realistically simulate pedestrians in diverse driving scenarios is key to evaluating the reliability of autonomous driving systems at scale. While recent advances in 3D Gaussian Splatting (3DGS) have enabled high-fidelity reconstruction of driving scenes, accurately representing pedestrians remains challenging. Existing annotation-free reconstruction methods lack explicit, posable pedestrian representations, and methods that do model pedestrians using parametric bodies rely on precise annotated 3D trajectories. Building on IDSplat, we present \modelname{}, a zero-shot 3DGS framework for reconstructing instance-decomposed dynamic driving scenes with temporally consistent, posable pedestrian representations. For each detected pedestrian, we leverage \sambody{} to extract per-frame initial body poses and initialize its Gaussians in the body model frame, ensuring they move naturally with body articulation. 
Through scene optimization, our framework jointly refines pedestrian body shapes, dynamic poses, and overall scene appearance. With this representation, we can repose pedestrians independently of their global trajectories and generate realistic counterfactual scenarios. In addition, we propose a lightweight, learnable pose interpolant to recover smooth and natural pedestrian poses at unobserved timestamps. Experiments on several automotive datasets show better reconstruction and novel view synthesis, particularly for pedestrians, than annotation-free baselines and even some methods using ground-truth 3D boxes. Code will be released.
\end{abstract}    
\section{Introduction}
\label{sec:intro}
Recent advances in 3D Gaussian Splatting (3DGS) have enabled the reconstruction and modification of driving scenes from collected sensor data~\cite{hess2025splatad,chen2025omnire,peng2025desire,xu2025ad,lindstrom2026idsplat,mao2026unire}, enabling the testing of autonomous driving (AD) systems in realistic simulated scenarios \cite{zhou2026hugsim}.
To account for moving road users, scenes are represented using time-varying primitives \cite{peng2025desire,chen2026periodic} or by having a global representation for the static environment and separate (local) representations for individual dynamic objects~\cite{yan2024streetgs,zhou2024hugs,zhou2024drivinggaussian,mao2026unire,lindstrom2026idsplat}. The latter lets objects move independently of the background, enabling trajectory modifications and the simulation of new traffic scenarios where road users can react to the AD system. 

Critically, the choice of object representation affects how we can manipulate the scene as well as how well we can reconstruct it, especially under sparse viewpoints and without annotated 3D bounding boxes.
For example, UnIRe~\cite{mao2026unire} groups observations of moving objects into instances and learns how the Gaussians within each instance deform over time.
This allows different parts of an object to move independently, but the flexibility comes at the cost of difficulty learning the true motion when only a few training views are available.
IDSplat~\cite{lindstrom2026idsplat} instead represents each dynamic object using a shared set of Gaussians and learns a joint rigid transformation at each time step. 
Clearly, this provides a stronger geometric and appearance regularization on the learned representation, which suits rigid objects such as vehicles and helps learn them from sparse views. 
However, it fails for pedestrians where, \eg, arms and legs move relative to the torso. As a result, for moving pedestrians, only the upper body is reconstructed well, while the arms and legs get blurry or are simply not reconstructed at all. 

An animatable representation~\cite{loper2015smpl, osman2020star, ferguson2025mhr, xu2020ghumghuml} can instead model the pedestrian's full movement and, thus, enable the model to represent and reconstruct both the global and local movement of the pedestrian. 
OmniRe~\cite{chen2025omnire} demonstrates this by representing pedestrians with the parametric Skinned Multi-Person Linear Model (SMPL)~\cite{loper2015smpl} and binding the Gaussians to the body model so they follow its pose. 
It uses ground-truth 3D tracklets to obtain pedestrian trajectories and to associate per-frame SMPL estimates from 4D-Humans~\cite{goel20234dhumans} with each pedestrian.
Although this allows pedestrians to be reconstructed with changing body poses, the method depends on annotated object tracks, which limits its applicability to unannotated driving logs and its scalability to other datasets. 

In this work, we present \modelname{}, which extends the zero-shot reconstruction method IDSplat~\cite{lindstrom2026idsplat} to handle pedestrians with changing body poses. We use 
the pretrained \sambody{} model~\cite{yang2026sam3dbody}, which predicts a body pose, parametrized by the Momentum Human Rig (\mhr{})~\cite{ferguson2025mhr} model, from each pedestrian detection. 
However, zero-shot segmentation and tracking can produce inaccurate masks and inconsistent associations, which hinders reliable pedestrian modeling and reconstruction. Furthermore, body shapes and poses predicted independently from individual images can vary across frames, even for the same person. We address these challenges by refining our tracked instance masks, learning a consistent body shape for each pedestrian, and jointly refining its pose and appearance during scene optimization; see \cref{fig:topfigure}. 
Furthermore, to better reconstruct pedestrians at held-out frames, we propose a pose interpolation model that combines interpolation of joint rotations with a learned correction network. The network refines the interpolated poses to more faithfully capture pedestrian movements, particularly at lower training frame rates. In experiments on two urban driving datasets with moving pedestrians, the Waymo Open Dataset~\cite{sun2020waymo} and PandaSet~\cite{xiao2021pandaset}, \modelname{} achieves the best novel view synthesis performance of all compared methods, including some that use annotations, with the largest gains on pedestrians.

Finally, since one use case of these reconstructions is testing AD systems in simulation, beyond image quality metrics, we evaluate the real-to-sim gap for pedestrians, \ie, the difference in how perception models trained on real data respond to reconstructed and real images.
Pretrained detection and body pose estimation models show a smaller gap for our method than for rigid or freely deforming instance representations. The learned pose interpolation further reduces this gap when training frames are sparse. These results indicate that modeling pedestrian body motion improves the realism of reconstructed pedestrians according to downstream perception models. 

In summary, our contributions are as follows.
\begin{itemize}
  \item A robust zero-shot method that learns a consistent posable body model for each pedestrian in driving scenes from noisy per-frame pose estimates and tracks, without annotated 3D bounding boxes.
  \item A pose interpolation model with a learned correction network that generates plausible body motion at held-out frames, in particular when training views are sparse.
  \item State-of-the-art novel view synthesis among annotation-free methods on Waymo and PandaSet, and a smaller real-to-sim gap for pedestrians in downstream pedestrian detection and pose estimation.
\end{itemize}
\section{Related Works}
\label{sec:related}

\parsection{Dynamic driving scene reconstruction}
Neural radiance fields (NeRFs)~\cite{mildenhall2020nerf} and 3DGS~\cite{kerbl20233dgs} have made photorealistic reconstruction of driving environments from recorded images~\cite{tancik2022blocknerf,tonderski2024neurad,hess2025splatad, zhou2024drivinggaussian, chen2025omnire, lindstrom2026idsplat, chen2025snerfpp} possible.
To handle moving objects, many methods reconstruct the static background separately and use ground-truth 3D bounding boxes to determine the pose of each dynamic actor over time~\cite{yan2024streetgs,zhou2024hugs,zhou2024drivinggaussian, hess2025splatad, ma2025beziergs}.
Methods that avoid annotated boxes instead identify dynamic regions directly from data to learn their movement over time~\cite{yang2023emernerf,chen2026periodic,peng2025desire,xu2025ad,wei2025emd}, but these methods distinguish moving regions in the scene without explicitly tracking individual objects. Instance-decomposed approaches, however, provide separate instances for dynamic actors~\cite{mao2026unire,lindstrom2026idsplat,su2025dial,tourani2025sdf}.
These instances can be moved using a single rigid transformation, as in IDSplat~\cite{lindstrom2026idsplat}, or represented with time-dependent deformation. For example, UnIRe~\cite{mao2026unire} decomposes the scene into instances using lidar and learns their motion relative to a canonical representation from estimated scene flow.
These approaches reconstruct moving actors without annotated boxes, but they do not represent pedestrian motion with a parametric body model that allows editing individual joints.

\parsection{Human body models and mesh recovery}
Parametric body models describe human shape and pose using a skeleton and a surface mesh~\cite{loper2015smpl, osman2020star, xu2020ghumghuml, ferguson2025mhr}, and pretrained estimators recover their parameters from single images~\cite{goel20234dhumans,yang2026sam3dbody}. The mesh is posed with linear blend skinning (LBS)~\cite{magnenatthalmann1988lbs,lewis2000psd}, or related skinning methods~\cite{kavan2007dqs}, which deform each vertex according to a weighted combination of the joint transformations.

\parsection{Pedestrian modeling}
Gaussian avatar methods attach Gaussians to parametric body models, placing them on or near the mesh surface and moving them with the body pose~\cite{hu2024gauhuman,qian20243dgsavatar,lei2024gart,moon2024exavatar,shao2024splattingavatar}. Similarly, OmniRe~\cite{chen2025omnire} represents pedestrians in driving scenes with SMPL~\cite{loper2015smpl} and binds Gaussians to the body mesh so that they follow its pose. \modelname{} follows a similar approach, but instead of requiring ground-truth pedestrian tracklets, we estimate pedestrian trajectories with zero-shot instance decomposition and use \sambody{}~\cite{yang2026sam3dbody} to estimate \mhr{} parameters, since it can be conditioned on the 2D instance masks from this decomposition, which ties each body estimate to a single pedestrian. Unlike SMPL, \mhr{} parametrizes the skeleton separately from the surface shape, so each pedestrian keeps a single set of bone lengths across frames. We refine these bone lengths together with the body poses and pedestrian appearance to learn a single re-poseable representation.
Animating a body requires interpolating between observed poses. Learned in-betweening methods~\cite{harvey2020inbetweening,oreshkin2024deltainterpolator} and generative motion models~\cite{rempe2021humor,tevet2023mdm,cohan2024condmdi} can predict intermediate poses from the surrounding observed ones, but are trained on clean, densely sampled motion capture data, and most are defined on the SMPL skeleton. In driving logs, poses are instead single-image estimates at the sensor frame rate. We therefore learn to interpolate poses from single-image body estimates. Combined with the instance decomposition of IDSplat~\cite{lindstrom2026idsplat}, the body model allows us to reconstruct and animate individual pedestrians within a complete driving scene, without annotations.
\vspace{-1mm}

\section{Method}
\label{sec:method}

\begin{figure*}[t]
  \centering
  \includegraphics[width=0.95\textwidth]{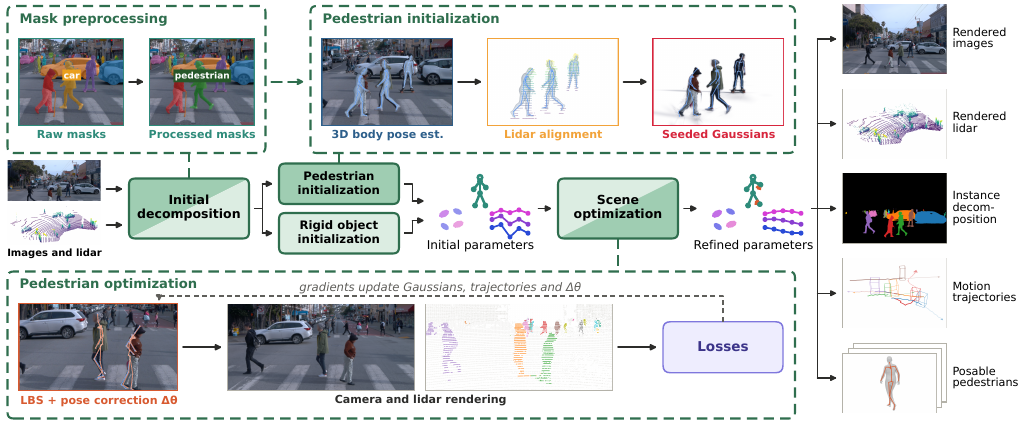}
  \caption{Overview of \modelname{}. Dark green marks our additions, light green the components inherited from IDSplat~\cite{lindstrom2026idsplat}, and split boxes both. After mask preprocessing, instances and their trajectories are estimated from the tracked masks and lidar. For each pedestrian, \sambody{} estimates an \mhr{} body from the training images, which is aligned with the lidar and used to initialize Gaussians. During scene optimization, the Gaussians move with the body through LBS, and per-frame pose corrections are refined with rendering losses.}
  \label{fig:overview}
  \vspace{-3mm}
\end{figure*}

Our goal is to reconstruct dynamic driving scenes from images and lidar point clouds, with independently moving objects and temporally consistent, posable pedestrian representations, without 3D annotations. We begin by reviewing the relevant background on 3DGS, IDSplat~\cite{lindstrom2026idsplat}, and parametric human body modeling (\cref{sec:prelim}). We then describe how we obtain and refine tracked instance masks (\cref{sec:sam3}), followed by pedestrian body estimation (\cref{sec:body_est}) and Gaussian initialization (\cref{sec:pedestrian_init}). Finally, we explain how pedestrian Gaussians are transformed according to body poses and global trajectories, how body parameters are refined, and how poses are interpolated at held-out frames (\cref{sec:motion}). \Cref{fig:overview} provides an overview of the method. 

\subsection{Preliminaries}
\label{sec:prelim}

\parsection{Scene representation} The scene is represented by translucent 3D Gaussians~\cite{kerbl20233dgs}, each parameterized by a mean $\boldsymbol{\mu}\in\mathbb{R}^3$, covariance $\mathbf{\Sigma}\in\mathbb{R}^{3\times3}$, and opacity $o\in[0,1]$, to describe its position, shape, and visibility.
Following SplatAD~\cite{hess2025splatad} and IDSplat~\cite{lindstrom2026idsplat}, we represent Gaussian appearance using feature vectors for base color and view-dependent effects, together with lidar properties. Images and lidar data are rendered following these methods.

\parsection{IDSplat} IDSplat~\cite{lindstrom2026idsplat} decomposes driving scenes into a static background and independently moving object instances. Given tracked 2D masks from Grounded-SAM-2~\cite{ren2024groundedsam2}, it associates lidar points with each object and estimates initial trajectories from DINOv3~\cite{simeoni2025dinov3} feature correspondences. Each instance, including pedestrians, is assumed to be rigid. The trajectories are refined by iterative coordinated turn smoothing and further optimized through rendering losses. \modelname{} follows this approach but models pedestrians as posable bodies, adding the following to the pipeline (\cref{fig:overview}). The masks are obtained with SAM~3~\cite{carion2026sam3} and refined (\cref{sec:sam3}) to improve instance tracking errors. Pedestrians are initialized from estimated \mhr{} bodies (\cref{sec:body_est,sec:pedestrian_init}). During optimization, the skeleton and per-frame poses of each pedestrian are refined with the rendering losses, and poses at held-out frames are predicted by a learned interpolation (\cref{sec:motion}).

\parsection{\mhr{} and \sambody} \mhr{}~\cite{ferguson2025mhr} describes the human body using a skeleton with $n_j=127$ joints and a surface mesh of $n_v=18,439$ vertices at the highest level of detail. The model has $n_{pose}=136$ pose parameters $\boldsymbol{\theta}$ controlling joint rotations, and $n_{skel}=68$ skeleton parameters $\boldsymbol{\sigma}$ defining bone lengths and other skeletal proportions. 
Given $\boldsymbol{\theta}$ and $\boldsymbol{\sigma}$, \mhr{} computes a transformation $\mathbf{T}^j$ for each joint $j$, describing its position and orientation relative to the model's coordinate system, which we call the body frame or the body coordinate system here. LBS~\cite{magnenatthalmann1988lbs} deforms the surface mesh using the joint transformations $\mathbf{T}^j$ and fixed weights $w^{ij}$ describing the influence of joint $j$ on vertex $i$.
\sambody{}~\cite{yang2026sam3dbody} predicts the \mhr{} parameters and the body translation relative to the camera from a single image, using a person mask as a prompt. 
\vspace{-1mm}
\subsection{Instance mask pre-processing and tracking}
\label{sec:sam3}
We use SAM~3~\cite{carion2026sam3} to detect and track 2D instance masks in each camera, prompted with text categories covering vehicles, cyclists, and pedestrians. Before associating the masklets with lidar points, we refine them to improve track consistency and resolve duplicate detections. A masklet is split into separate instances when the object reappears with an inconsistent bounding box after at least four frames. Duplicate masklets are merged by overlap, keeping the most confident class label. Split segments of an object that no other masklet covers are attached to an adjacent instance when most of their pixels border its masklet in shared frames, so that associated objects such as backpacks are rendered as part of the pedestrian. We then estimate the trajectory of each object, its pose $\mathbf{P}_t$ at each time $t$, from the refined masks and lidar points, following IDSplat.
IDSplat assigns stationary pedestrians to the static background. However, a pedestrian who stays in place can still move their limbs. We therefore keep stationary pedestrians as separate actors when a valid body estimate is available.
\vspace{-1mm}
\subsection{Pedestrian body estimation}
\label{sec:body_est}

We apply \sambody{} to each person in the training images, prompted with the refined SAM~3 masklets, so each predicted body is associated with an instance and its lidar points. We align the mesh with the lidar points by a translation estimated with ICP~\cite{besl1992icp}. Since \sambody{} estimates each frame independently, body shape and bone lengths vary across the frames of a pedestrian. We therefore initialize one set of skeleton parameters $\boldsymbol{\sigma}_0$ per pedestrian as their median over the track. These are refined during scene optimization, while the remaining pose parameters, the global rotation and joint angles, are optimized per frame.

\subsection{Pedestrian Gaussian representation}
\label{sec:pedestrian_init}

\parsection{Canonical frame}
The canonical frame $c$ is the frame with the most lidar points on the pedestrian or, if it has no valid body estimate, the first frame with one. \sambody{} outputs each body in a body frame with its origin near the pelvis and the axes of the observing camera. From the canonical instance pose $\mathbf{P}_c$, the camera pose and the predicted body translation at $c$, we compute a fixed transform $\mathbf{M}$ from the instance’s local frame to the body frame. 
The pedestrian Gaussians are parameterized in the instance’s local frame. To apply a pose, they are mapped to the body frame with $\mathbf{M}$, deformed by LBS, and mapped back, and the trajectory then places the local frame in the scene.

\parsection{Gaussian initialization}
For each pedestrian, we initialize $9220$ Gaussians (every second vertex of the \mhr{} mesh) in the canonical frame. Their positions are transformed into the local frame using $\mathbf{M}^{-1}$. Each Gaussian keeps the index $i$ of its source vertex, which determines its skinning weights $w^{ij}$.
The estimated body mesh does not explicitly capture loose clothing, hair, or carried objects, and may not align perfectly with the pedestrian surface. To represent geometry beyond the parametric body model, we also initialize Gaussians from the pedestrian's lidar points. We use only points from the canonical frame, since accumulating lidar points across frames could introduce inconsistent geometry due to changes in body pose. Each lidar Gaussian inherits the skinning weights of its nearest body Gaussian to follow the motion of the corresponding body part.

\subsection{Pedestrian motion}
\label{sec:motion}

Pedestrian motion consists of the global movement of the instance throughout the scene and changes in the body pose.

\parsection{Body deformation}
Each pedestrian Gaussian is moved in three steps. Its local mean $\boldsymbol{\mu}$ is first mapped to the canonical frame, $\mathbf{x}_c = \mathbf{M}\boldsymbol{\mu}$. Then, LBS moves this point from the canonical pose to the pose at time $t$,
\begin{equation}
  \mathbf{x}_t = \Big(\sum_j w^{ij}\,\mathbf{T}^j_t\,(\mathbf{T}^j_c)^{-1}\Big)\,\mathbf{x}_c,
  \label{eq:lbs}
\end{equation}
where $i$ is the source vertex of the Gaussian and $\mathbf{T}_t^j({\mathbf{T}^j_c})^{-1}$ is the motion of joint $j$ since the canonical frame. Finally, the posed point is mapped back to the instance's local frame and placed in the scene, $\boldsymbol{\mu}_t = \tilde{\mathbf{P}}_t\,\mathbf{M}^{-1}\mathbf{x}_t$, where $\tilde{\mathbf{P}}_t$ is the instance pose at time $t$, obtained from its trajectory $\mathbf{P}_t$. 

\parsection{Body refinement}
Since \sambody{} estimates each frame independently, the initial body estimates may be inconsistent across frames and misaligned with the images and lidar points. We therefore refine the body parameters jointly with the Gaussians to obtain a consistent body representation for each pedestrian. Specifically, we optimize a shared offset to the skeleton parameters for each pedestrian and per-frame offsets $\Delta\boldsymbol{\theta}$ to the pose parameters. All offsets are initialized to zero and optimized using the rendering losses.
\sambody{} does not predict the 68 skeleton parameters directly. It predicts 28 coefficients $\mathbf{z}$, which are mapped to the skeleton parameters as $\boldsymbol{\sigma} = \bar{\boldsymbol{\sigma}} + \mathbf{B}\mathbf{z}$, where the mean $\bar{\boldsymbol{\sigma}}\in\mathbb{R}^{68}$ and the matrix $\mathbf{B}\in\mathbb{R}^{68\times 28}$ are fixed parts of the model.
We refine the skeleton in the same space: we optimize an offset $\Delta\mathbf{z}$ to the coefficients, which changes the skeleton parameters by $\mathbf{B}\Delta\mathbf{z}$ and gives $\boldsymbol{\sigma} = \boldsymbol{\sigma}_0 + \mathbf{B}\Delta\mathbf{z}$, where $\boldsymbol{\sigma}_0$ is the per-pedestrian initialization from \cref{sec:body_est}.
The joint transformations $\mathbf{T}^j_c$ at the canonical frame in \cref{eq:lbs} are computed once from the initial body, so the deltas $\mathbf{T}^j_t(\mathbf{T}^j_c)^{-1}$ include skeleton changes and the Gaussians follow the refined joints.

\parsection{Pose interpolation}
For novel-view synthesis where we have held-out frames without body parameters, we estimate pedestrian poses using a learned interpolation method. We compute an initial pose from the two observations surrounding the target time $t$. Joint angles are interpolated linearly along the shortest angular path, and global rotations are interpolated using spherical linear interpolation (SLERP)~\cite{shoemake1985slerp}. \mhr{} computes the joint transformations from these pose parameters and the pedestrian's skeleton, which ensures that bone lengths remain consistent. However, the resulting pose may not accurately reproduce realistic movements. Thus, a learned correction network adjusts the interpolated joint angles to better represent pedestrian movements between observations.

The correction network is a multilayer perceptron (MLP) trained on \sambody{} predictions of pedestrians in driving scenes that are excluded from the evaluation scenes. For each pedestrian, we withhold individual frames and train the network to recover them from the surrounding predictions.
Given the four nearest observations, two before and two after the target time $t$, the MLP predicts adjustments to the interpolated joint angles. 
Each observation is represented by its joint angles, its time offset relative to $t$, and its global rotation relative to the last observation before $t$. We minimize the mean distance between predicted and withheld joint positions, computed by \mhr{}. 

\section{Experiments}
\label{sec:experiments}

We evaluate whether a consistent, posable representation of each pedestrian improves reconstruction and novel view synthesis (NVS) of dynamic driving scenes without annotated 3D tracks. We also analyze the individual components, evaluate our learned interpolant with fewer training frames, and test the reconstructed pedestrians for downstream perception.

\subsection{Experimental setup}
\label{sec:setup}

We evaluate on the Waymo Open Dataset (WOD)~\cite{sun2020waymo}, whose urban scenes contain many moving pedestrians, and on PandaSet~\cite{xiao2021pandaset} to test other sensors and environments. On WOD, we follow the eight-sequence setting of OmniRe~\cite{chen2025omnire} and UnIRe~\cite{mao2026unire}, with three front cameras at half resolution and every tenth frame held out for NVS. We also report results with the front camera only, for comparison with AD-GS~\cite{xu2025ad}, and at full resolution, to aid future comparisons in higher-fidelity reconstructions on this dataset. On PandaSet, we follow IDSplat~\cite{lindstrom2026idsplat}, with all six cameras at full resolution and every second frame held out. Image reconstruction is evaluated on the training frames. For held-out frames, instance trajectories are interpolated between neighboring training frames and body poses are estimated with the learned interpolation of \cref{sec:motion}. We report PSNR, SSIM~\cite{wang2004ssim} and LPIPS~\cite{zhang2018lpips} on the full image, and the PSNR in dynamic regions (DPSNR)~\cite{yang2023emernerf} for pedestrians and vehicles. Lidar metrics follow SplatAD~\cite{hess2025splatad}. Our primary baseline is IDSplat, which we extend with posable pedestrians. IDSplat, SplatAD, DIAL-GS~\cite{su2025dial}, AD-GS and UnIRe are reproduced with their public implementations, and the remaining results are taken from their publications. More information on the evaluation protocol is presented in \cref{sec:supmat_dataeval}. Further evaluations on KITTI~\cite{geiger2012kitti,liao2021kitti360} and other WOD scenes, following CoDa-4DGS~\cite{song2025coda} and SplatFlow~\cite{sun2025splatflow}, are given in \cref{sec:supmat_results}.

\subsection{Implementation}
Our method is built on the IDSplat~\cite{lindstrom2026idsplat} code. The tracked 2D instance masks are generated with SAM~3\cite{carion2026sam3} and the \mhr{} bodies with \sambody{}\cite{yang2026sam3dbody}. Each scene is trained for 30{,}000 steps on one NVIDIA A100 GPU. Further implementation details are given in \cref{sec:supmat_impl} and the runtime analysis is provided in \cref{sec:supmat_runtime}.

\subsection{Baseline comparisons}
\label{sec:results}

Our experiments examine whether a consistent body model improves rendering at both observed and held-out frames. 

\begin{table*}[t]
  \centering
  \caption{Reconstruction and NVS on the OmniRe split of Waymo~\cite{sun2020waymo}. PSNR and DPSNR in dB.
    The full-res row is not comparable with the other rows.
    Colors rank the annotation-free methods; the front-camera rows are ranked between themselves. \colorbox{tabfirst}{First}, \colorbox{tabsecond}{second}, \colorbox{tabthird}{third}.}
    \vspace{-2mm}
  \label{tab:omnire}
  \footnotesize
  \setlength{\tabcolsep}{1.5pt}
  \begin{tabular}{@{}l c c ccccc @{\hskip 8pt} ccccc@{}}
    \toprule
    & & & \multicolumn{5}{c}{Reconstruction} & \multicolumn{5}{c}{Novel view synthesis} \\
    \cmidrule(lr){4-8} \cmidrule(l){9-13}
    Method & \makecell{Anno.\\free} & \makecell{Inst.\\decomp.} & PSNR$\uparrow$ & SSIM$\uparrow$ & LPIPS$\downarrow$ & \makecell{DPSNR$\uparrow$\\ped.} & \makecell{DPSNR$\uparrow$\\veh.}
                                 & PSNR$\uparrow$ & SSIM$\uparrow$ & LPIPS$\downarrow$ & \makecell{DPSNR$\uparrow$\\ped.} & \makecell{DPSNR$\uparrow$\\veh.} \\
    \midrule
    OmniRe~\cite{chen2025omnire}            & \xmark & \cmark & 34.81 & 0.956 & 0.054 & 27.56 & 28.91 & 33.03 & 0.944 & 0.060 & 24.20 & 27.78 \\
    SplatAD~\cite{hess2025splatad}          & \xmark & \cmark & 30.45 & 0.949 & 0.069 & 22.41 & 28.27 & 29.71 & 0.941 & 0.073 & 21.59 & 27.32 \\
    UnIRe w/ GT box~\cite{mao2026unire}     & \xmark & \cmark & 36.09 & 0.963 & 0.053 & 33.23 & 29.59 & 33.05 & 0.943 & 0.065 & 24.32 & 27.92 \\
    \dashrule{1-13}
    AD-GS~\cite{xu2025ad}, front cam.       & \cmark & \xmark & \cellcolor{tabsecond}38.78 & \cellcolor{tabfirst}0.977 & \cellcolor{tabsecond}0.045 & \cellcolor{tabsecond}35.03 & \cellcolor{tabfirst}35.03 & \cellcolor{tabsecond}33.95 & \cellcolor{tabsecond}0.956 & \cellcolor{tabsecond}0.060 & \cellcolor{tabsecond}27.66 & \cellcolor{tabsecond}29.14 \\
    \modelname{} (ours), front cam.                       & \cmark & \cmark & \cellcolor{tabfirst}39.33 & \cellcolor{tabfirst}0.977 & \cellcolor{tabfirst}0.025 & \cellcolor{tabfirst}35.81 & \cellcolor{tabsecond}34.69 & \cellcolor{tabfirst}35.03 & \cellcolor{tabfirst}0.964 & \cellcolor{tabfirst}0.033 & \cellcolor{tabfirst}28.32 & \cellcolor{tabfirst}30.29 \\
    \dashrule{1-13}
    EmerNeRF~\cite{yang2023emernerf}        & \cmark & \xmark & 31.51 & 0.891 & 0.112 & 22.73 & 24.76 & 29.53 & 0.878 & 0.139 & 21.37 & 21.98 \\
    PVG~\cite{chen2026periodic}             & \cmark & \xmark & 32.61 & 0.936 & 0.103 & 24.72 & 24.29 & 28.94 & 0.881 & 0.127 & 21.92 & 21.59 \\
    DeSiRe-GS~\cite{peng2025desire}         & \cmark & \xmark & 32.71 & 0.949 & 0.103 & 24.87 & 24.51 & 30.67 & 0.933 & 0.118 & 22.53 & 22.70 \\
    UnIRe~\cite{mao2026unire}               & \cmark & \cmark & \cellcolor{tabsecond}35.58 & \cellcolor{tabsecond}0.967 & \cellcolor{tabthird}0.053 & \cellcolor{tabsecond}30.44 & \cellcolor{tabsecond}30.62 & 31.56 & 0.935 & \cellcolor{tabthird}0.074 & 22.75 & 24.82 \\
    DIAL-GS~\cite{su2025dial}               & \cmark & \cmark & \cellcolor{tabthird}34.57 & 0.954 & 0.072 & \cellcolor{tabthird}30.15 & 29.09 & \cellcolor{tabthird}31.85 & \cellcolor{tabthird}0.936 & 0.087 & \cellcolor{tabsecond}26.20 & \cellcolor{tabthird}25.42 \\
    IDSplat~\cite{lindstrom2026idsplat}     & \cmark & \cmark & 34.25 & \cellcolor{tabthird}0.963 & \cellcolor{tabsecond}0.046 & 27.73 & \cellcolor{tabthird}30.39 & \cellcolor{tabsecond}32.46 & \cellcolor{tabsecond}0.951 & \cellcolor{tabsecond}0.052 & \cellcolor{tabthird}25.50 & \cellcolor{tabsecond}28.26 \\
    \modelname{} (ours)                                   & \cmark & \cmark & \cellcolor{tabfirst}36.69 & \cellcolor{tabfirst}0.971 & \cellcolor{tabfirst}0.035 & \cellcolor{tabfirst}32.31 & \cellcolor{tabfirst}31.79 & \cellcolor{tabfirst}33.95 & \cellcolor{tabfirst}0.957 & \cellcolor{tabfirst}0.041 & \cellcolor{tabfirst}27.60 & \cellcolor{tabfirst}29.07 \\
    \dashrule{1-13}
    \modelname{} (ours) full-res                                   & \cmark & \cmark & 34.62 & 0.951 & 0.087 & 31.49 & 30.49 & 32.55 & 0.940 & 0.093 & 27.26 & 28.16 \\
    \bottomrule
  \end{tabular}
\end{table*}

\begin{figure*}[t]
  \centering
  \includegraphics[width=0.88\textwidth]{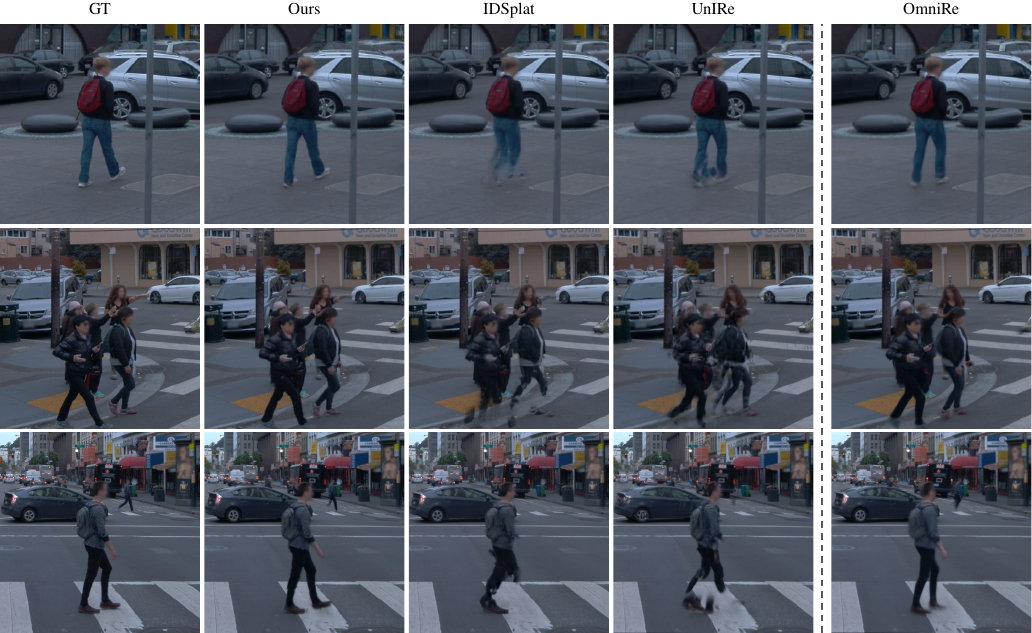}
  \caption{NVS of pedestrians at held-out frames of the WOD OmniRe split. 
  OmniRe uses ground-truth boxes and is shown separately.
  }
  \label{fig:quality}
  \vspace{-2mm}
\end{figure*}

\parsection{WOD}
\Cref{tab:omnire} shows that \modelname{} achieves the best reconstruction and NVS among annotation-free methods, and remains competitive with methods using ground-truth boxes. The same holds for a single camera in the comparison with AD-GS.
The comparison with IDSplat isolates the effect of pedestrian body motion. Our pedestrian representation, consistent across limb configurations, mainly improves pedestrian rendering at held-out timestamps, with little effect on vehicles, which keep the IDSplat representation.
UnIRe reconstructs observed frames accurately but degrades substantially more on held-out frames. This suggests that its flexible deformation fits observed frames well but generalizes poorly to unobserved poses. Our shared skeleton and Gaussians across frames instead constrain the appearance at unobserved poses.
\Cref{fig:quality} shows walking pedestrians at held-out frames. IDSplat and UnIRe render the legs smeared or broken. OmniRe, with annotated tracks and SMPL poses, renders separated legs, but its poses at held-out frames deviate from the ground truth. Our method renders separated legs at poses closer to the ground truth, which we attribute to the learned pose interpolation. More qualitative results are included in \cref{sec:supmat_qualitative}.

\begin{table}[t]
  \centering
  \caption{Lidar NVS (OmniRe split). Depth and CD in m$^2$. SplatAD has no pedestrian actors. Colors rank the annotation-free methods. \colorbox{tabfirst}{First}, \colorbox{tabsecond}{second}.}
  \vspace{-3mm}
  \label{tab:lidar}
  \footnotesize
  \setlength{\tabcolsep}{1.0pt}
  \begin{tabular}{@{}l cccc @{\hskip 6pt} cccc@{}}
    \toprule
    & \multicolumn{4}{c}{All points} & \multicolumn{4}{c}{Pedestrian points} \\
    \cmidrule(lr){2-5} \cmidrule(l){6-9}
    Method & Depth$\downarrow$ & Int.$\downarrow$ & Drop$\uparrow$ & CD$\downarrow$ & Depth$\downarrow$ & Int.$\downarrow$ & Drop$\uparrow$ & CD$\downarrow$ \\
    \midrule
    SplatAD~\cite{hess2025splatad}              & 0.008 & 0.060 & 90.0 & 1.13 & -- & -- & -- & -- \\
    \dashrule{1-9}
    IDSplat~\cite{lindstrom2026idsplat}         & \cellcolor{tabsecond}0.007 & \cellcolor{tabsecond}0.058 & \cellcolor{tabsecond}89.8 & \cellcolor{tabsecond}1.26 & \cellcolor{tabsecond}0.073 & \cellcolor{tabsecond}0.093 & \cellcolor{tabsecond}97.6 & \cellcolor{tabsecond}0.20 \\
    \modelname{} (ours)                                       & \cellcolor{tabfirst}0.005 & \cellcolor{tabfirst}0.056 & \cellcolor{tabfirst}89.9 & \cellcolor{tabfirst}1.14 & \cellcolor{tabfirst}0.010 & \cellcolor{tabfirst}0.079 & \cellcolor{tabfirst}97.9 & \cellcolor{tabfirst}0.10 \\
    \bottomrule
  \end{tabular}
\end{table}

\parsection{Lidar rendering}
\Cref{tab:lidar} evaluates whether the posable body also improves pedestrian geometry and lidar rendering. For pedestrians, \modelname{} outperforms IDSplat at held-out sweeps in all metrics, showing that the body model places the pedestrian surface more accurately, not only its appearance in the images. The selection of pedestrian boxes is detailed in \cref{sec:supmat_dataeval}. On all points, our method is on par with SplatAD, which uses ground-truth boxes, and IDSplat, with which we share the background and vehicle representations.

\begin{table}[t]
  \centering
  \caption{NVS on PandaSet~\cite{xiao2021pandaset}. Colors rank the annotation-free methods. \colorbox{tabfirst}{First}, \colorbox{tabsecond}{second}, \colorbox{tabthird}{third}.}
  \vspace{-3mm}
  \label{tab:pandaset}
  \footnotesize
  \setlength{\tabcolsep}{2pt}
  \begin{tabular}{@{}l c c cccc@{}}
    \toprule
    Method & \makecell{Anno.\\free} & \makecell{Inst.\\decomp.} & PSNR$\uparrow$ & SSIM$\uparrow$ & LPIPS$\downarrow$ & \makecell{DPSNR$\uparrow$\\ped.} \\
    \midrule
    StreetGS~\cite{yan2024streetgs}       & \xmark & \cmark & 24.73 & 0.745 & 0.314 & -- \\
    SplatAD~\cite{hess2025splatad}        & \xmark & \cmark & 26.76 & 0.815 & 0.193 & -- \\
    \dashrule{1-7}
    PVG~\cite{chen2026periodic}           & \cmark & \xmark & \cellcolor{tabthird}24.01 & \cellcolor{tabthird}0.712 & \cellcolor{tabthird}0.452 & -- \\
    IDSplat~\cite{lindstrom2026idsplat} & \cmark & \cmark & \cellcolor{tabsecond}26.78 & \cellcolor{tabsecond}0.814 & \cellcolor{tabfirst}0.174 & \cellcolor{tabsecond}23.11 \\
    \modelname{} (ours)                                 & \cmark & \cmark & \cellcolor{tabfirst}26.82 & \cellcolor{tabfirst}0.815 & \cellcolor{tabfirst}0.174 & \cellcolor{tabfirst}23.46 \\
    \bottomrule
  \end{tabular}
  \vspace{-3mm}
\end{table}

\parsection{PandaSet}
On PandaSet (\cref{tab:pandaset}), \modelname{} matches IDSplat on full-image metrics and improves pedestrian DPSNR, consistent with the small image area that pedestrians occupy in these scenes. 

\subsection{Pedestrian instances}
\label{sec:decomposition}
\begin{figure}[t]
  \centering
  \includegraphics[width=0.9\linewidth]{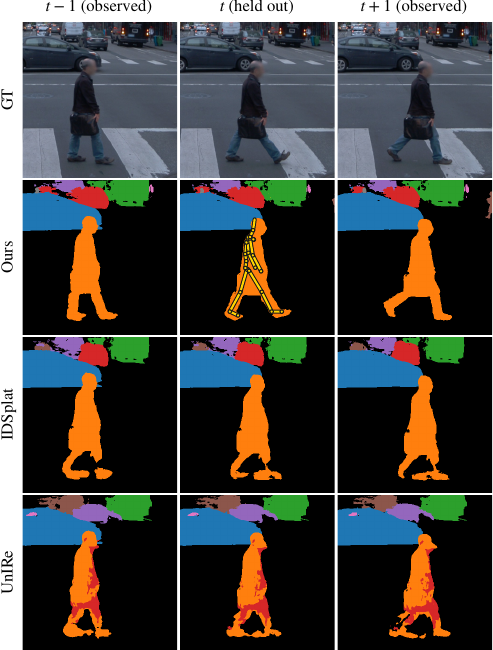}
  \caption{Rendered instance masks of a pedestrian at a held-out frame $t$ and at the neighboring training frames (OmniRe split), with our projected \mhr{} skeleton at $t$.
  }
  \label{fig:decomposition}
\end{figure}

\vspace{-2mm}
To be controllable, a pedestrian must be reconstructed as one instance whose shape follows its pose. \Cref{fig:decomposition} shows the rendered instance masks of a walking pedestrian at a held-out frame and the two neighboring training frames.  Our method renders the pedestrian as one instance whose legs follow the stride of the projected skeleton at the held-out frame. IDSplat also renders one instance, but its legs barely change across the three frames as the pedestrian moves forward rigidly, so it can be moved but not posed. UnIRe assigns part of the pedestrian to a second instance, so it cannot be edited as one object. This split does not prevent UnIRe from fitting the training frames, where it has the second highest pedestrian DPSNR among annotation-free methods (\cref{tab:omnire}), but at held-out frames it falls below DIAL-GS and IDSplat, suggesting that its freely deforming instances overfit the observed frames. More examples of instance decomposition performance are included in \cref{sec:supmat_qualitative}.

\subsection{Ablation study}
\label{sec:ablations}

\begin{figure}[t]
  \centering
  \includegraphics[width=0.98\linewidth]{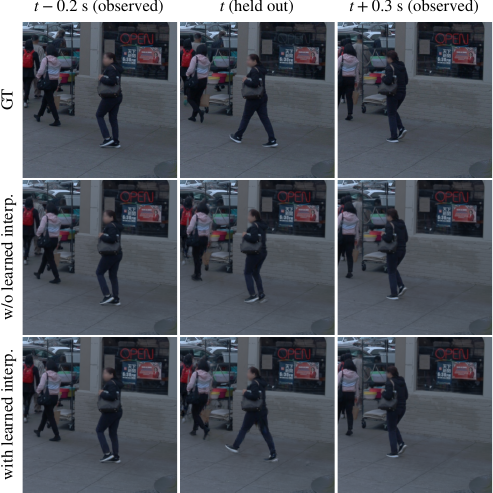}
  \caption{Learned interpolation with 25\% of the frames for training. At held-out frame $t$, interpolating the observed poses leaves the legs together, while learned correction recovers the stride.}
  \label{fig:prior}
  \vspace{-4mm}
\end{figure}

\Cref{tab:ablations} removes one component at a time and evaluates NVS on the OmniRe split. Since our method relies on zero-shot masks instead of ground-truth boxes, it must be robust to mask errors. Without preprocessing, tracks that drift between objects are not split and duplicate masklets are not resolved, so instances may mix objects or lose observations. These errors are not specific to pedestrians: vehicle DPSNR decreases even more. Removing LBS causes the largest drop in pedestrian DPSNR, showing the limitations of rigid pedestrians. Without body refinement, pedestrians keep the median skeleton and per-frame \sambody{} poses, and the drop shows the benefit of refining single-image body estimates with rendering losses. Lidar-initialized Gaussians give a smaller improvement. The correction network has no measurable effect when every tenth frame is held out, as most held-out frames lie between consecutive training frames, but its benefit grows with sparser training data. In \cref{fig:prior}, trained on every fourth frame (25\%), linear interpolation leaves the legs together while the correction network recovers the stride. The downstream evaluations in \cref{tab:downstream} and \cref{sec:supmat_interp} give further results regarding the learned interpolation approach. 

\begin{table}[t]
    \centering
    \caption{Ablations (NVS on the OmniRe split). Each row removes one component of our method.}
    \vspace{-3mm}
    \label{tab:ablations}
    \footnotesize
    \setlength{\tabcolsep}{2.0pt}
    \begin{tabular}{@{}lccccc@{}}
          \toprule
          & PSNR$\uparrow$ & SSIM$\uparrow$ & LPIPS$\downarrow$ & DPSNR\textsubscript{ped}$\uparrow$ & DPSNR\textsubscript{veh}$\uparrow$ \\
          \midrule
          \ours                  & 33.95 & 0.957 & 0.041 & 27.60 & 29.07 \\
          w/o mask preproc.      & 33.23 & 0.954 & 0.046 & 26.83 & 27.82 \\
          w/o LBS                & 33.14 & 0.954 & 0.048 & 26.30 & 29.02 \\
          w/o lidar Gaussians    & 33.89 & 0.957 & 0.042 & 27.39 & 29.13 \\
          w/o pose refinement    & 33.70 & 0.956 & 0.043 & 27.27 & 28.92 \\
          w/o learned interp.    & 33.96 & 0.957 & 0.041 & 27.60 & 29.12 \\
          \bottomrule
    \end{tabular}
    \vspace{-4mm}
\end{table}

\subsection{Downstream perception}
\label{sec:downstream}

To test whether our pedestrians are more useful for perception beyond image reconstruction metrics, we run pretrained object detectors and pose estimators on rendered held-out frames and compare their predictions with those on the corresponding real images.
We use ViTDet~\cite{li2022vitdet} and Grounding DINO~\cite{liu2024groundingdino} for pedestrian detection and ViTPose~\cite{xu2022vitpose} for keypoint estimation, without fine-tuning, on the OmniRe split (\cref{tab:downstream}). We exclude SAM~3 and \sambody{}, on which our method is based. Detection is measured by AP50~\cite{everingham2010pascal} against the WOD 2D boxes and by the detection agreement (DA) between rendered and real images~\cite{lindstrom2024real2sim}, and keypoints by PCK~\cite{yang2013articulated} and OKS~\cite{lin2014coco}. Further detail is available in \cref{sec:supmat_downstream}.

At every training density, the detectors and the pose estimator respond to our renderings more as they do to real images than to those of IDSplat and UnIRe. The gap is largest for pose estimation, which depends most directly on the limb configuration. UnIRe scores lowest on every metric, consistent with its degradation at held-out frames (\cref{tab:omnire}).
The correction network matters most where the gaps between observations are longest. With training on 50\% of the frames it has almost no effect, while with 25\% it improves keypoint accuracy and detection agreement without changing AP50, indicating that the corrected poses are also more plausible to a pose estimator not involved in training. Still, detection and pose estimation on our rendered images remain below the real images at every frame density.

\providecommand{\gap}[1]{\,{\scriptsize\color{gray}(#1)}}
  \begin{table}[t]
    \centering
    \caption{Pedestrian detection and pose estimation on rendered held-out frames of the OmniRe split, trained on 50\%, 33\% and 25\% of frames. AP50, DA, PCK in \%.
    \colorbox{tabfirst}{First}, \colorbox{tabsecond}{second}, \colorbox{tabthird}{third}.}
    \label{tab:downstream}
    \footnotesize
    \setlength{\tabcolsep}{1.0pt}
    \begin{tabular}{@{}cl cc cc cc@{}}
      \toprule
      & & \multicolumn{2}{c}{ViTDet} & \multicolumn{2}{c}{Grounding DINO} & \multicolumn{2}{c}{ViTPose} \\
      \cmidrule(lr){3-4} \cmidrule(lr){5-6} \cmidrule(l){7-8}
      & Images & AP50$\uparrow$ & DA$\uparrow$ & AP50$\uparrow$ & DA$\uparrow$ & PCK$\uparrow$ & OKS$\uparrow$ \\
      \midrule
      \multirow{5}{*}{\rotatebox[origin=c]{90}{\textit{50\% frames}}}
      & Real & 68.4 & -- & 75.1 & -- & 95.4 & 0.882 \\
      & UnIRe~\cite{mao2026unire} & 48.9 & 72.7 & \cellcolor{tabthird}53.1 & 77.3 & 72.7 & 0.643 \\
      & IDSplat~\cite{lindstrom2026idsplat} & \cellcolor{tabthird}51.2 & \cellcolor{tabthird}75.5 & \cellcolor{tabsecond}55.7 & \cellcolor{tabthird}79.3 & \cellcolor{tabthird}74.8 &
  \cellcolor{tabthird}0.664 \\
      & \ours{} w/o learned interp. & \cellcolor{tabfirst}56.8 & \cellcolor{tabsecond}82.5 & \cellcolor{tabfirst}62.1 & \cellcolor{tabsecond}86.8 & \cellcolor{tabfirst}84.4 & \cellcolor{tabfirst}0.755 \\
      & \ours & \cellcolor{tabsecond}56.7 & \cellcolor{tabfirst}82.7 & \cellcolor{tabfirst}62.1 & \cellcolor{tabfirst}87.2 & \cellcolor{tabsecond}84.3 & \cellcolor{tabsecond}0.754 \\
      \midrule
      \multirow{5}{*}{\rotatebox[origin=c]{90}{\textit{33\% frames}}}
      & Real & 68.3 & -- & 75.1 & -- & 95.2 & 0.880 \\
      & UnIRe~\cite{mao2026unire} & \cellcolor{tabthird}48.8 & 71.7 & \cellcolor{tabthird}53.5 & 76.7 & 72.7 & 0.642 \\
      & IDSplat~\cite{lindstrom2026idsplat} & \cellcolor{tabsecond}51.2 & \cellcolor{tabthird}74.9 & \cellcolor{tabsecond}56.3 & \cellcolor{tabthird}79.0 & \cellcolor{tabthird}74.9 &
  \cellcolor{tabthird}0.671 \\
      & \ours{} w/o learned interp. & \cellcolor{tabfirst}55.4 & \cellcolor{tabsecond}80.3 & \cellcolor{tabfirst}60.9 & \cellcolor{tabsecond}83.8 & \cellcolor{tabsecond}81.7 & \cellcolor{tabsecond}0.733 \\
      & \ours & \cellcolor{tabfirst}55.4 & \cellcolor{tabfirst}81.0 & \cellcolor{tabfirst}60.9 & \cellcolor{tabfirst}84.9 & \cellcolor{tabfirst}81.8 & \cellcolor{tabfirst}0.735 \\
      \midrule
      \multirow{5}{*}{\rotatebox[origin=c]{90}{\textit{25\% frames}}}
      & Real & 68.4 & -- & 75.0 & -- & 95.3 & 0.882 \\
      & UnIRe~\cite{mao2026unire} & 46.0 & 68.2 & \cellcolor{tabthird}50.3 & 73.2 & 69.7 & 0.615 \\
      & IDSplat~\cite{lindstrom2026idsplat} & \cellcolor{tabthird}50.4 & \cellcolor{tabthird}73.6 & \cellcolor{tabsecond}55.4 & \cellcolor{tabthird}78.0 & \cellcolor{tabthird}74.5 &
  \cellcolor{tabthird}0.665 \\
      & \ours{} w/o learned interp. & \cellcolor{tabsecond}54.0 & \cellcolor{tabsecond}77.6 & \cellcolor{tabfirst}59.8 & \cellcolor{tabsecond}81.8 & \cellcolor{tabsecond}80.5 & \cellcolor{tabsecond}0.721 \\
      & \ours & \cellcolor{tabfirst}54.2 & \cellcolor{tabfirst}79.1 & \cellcolor{tabfirst}59.8 & \cellcolor{tabfirst}83.7 & \cellcolor{tabfirst}81.2 & \cellcolor{tabfirst}0.727 \\
      \bottomrule
    \end{tabular}
    \vspace{-4mm}
  \end{table}

\subsection{Re-posing pedestrians}
\label{sec:reposing}

Beyond reproducing the recorded scene, a posable pedestrian representation allows body motion to be modified independently of the global instance trajectory. Since each pedestrian has a single skeleton and its Gaussians are attached to the body, it can be re-posed without retraining. \Cref{fig:topfigure} demonstrates this by transferring estimated poses between reconstructed pedestrians while preserving the target’s body shape and appearance. 

\subsection{Limitations}
\label{sec:limitations}
Our method reconstructs pedestrians with high fidelity and allows re-posing, but some limitations remain. First, the body model covers the body only, and accessories are not modeled explicitly. Carried objects and loose clothing follow the nearest body part, so they may move unnaturally when the body is re-posed. Second, a pedestrian is reconstructed only from the observed views, so edits are limited to poses and positions that show only observed body parts. Third, shadows and reflections that move with a dynamic object are not modeled explicitly and may be absorbed by its Gaussians, so they may stay attached to the object when it is moved instead of following the scene lighting. Finally, the method depends on zero-shot models. We refine the SAM~3 masklets before use and the \sambody{} estimates during optimization, but errors may occur: a pedestrian missed by SAM~3 is not reconstructed as an instance, and a body estimate that confuses the front and back of a pedestrian may produce a turn the pedestrian did not make unless the refinement corrects it. These errors are more likely for distant pedestrians. We discuss these limitations further in \cref{sec:supmat_limitations}.

\section{Conclusion}
\label{sec:conclusion}
We presented \modelname{}, an annotation-free method for reconstructing and animating pedestrians in driving scenes. Combining instance decomposition with pretrained human body estimation, we represent pedestrians with Gaussians initialized from the \mhr{} mesh and lidar points, animated through linear blend skinning. We refine body poses during scene optimization and use learned interpolation to better predict poses at held-out frames, especially with sparse training views. Experiments on the OmniRe split show improved pedestrian reconstruction over both annotation-free methods and methods using ground-truth 3D annotations. Moreover, pretrained object detectors and pose estimators are more accurate on our renders than on those of the baselines, especially for pose estimation.

\section*{Acknowledgments}
\label{sec:}
We thank Carl Lindström and Adam Lilja for valuable feedback. This work was partially supported by Vinnova, the Swedish Innovation Agency. Computational resources were provided on the Berzelius system funded by the Knut and Alice Wallenberg foundation and operated by NAISS.

\section*{LLM usage statement}
\label{sec:llm}
The Claude model from Anthropic was used to assist with language editing and restructuring of author-written text, refinement and optimization of author-written code, and preparation of figures and visualizations throughout the paper. The research ideas, methodology, experimental design, analysis, and conclusions were developed by the authors without LLM assistance. The authors reviewed and verified the final content and take full responsibility for the manuscript. 

{
    \small
    \bibliographystyle{ieeenat_fullname}
    \bibliography{main}

@String(CVPR= {IEEE Conf. Comput. Vis. Pattern Recog.})

@String(ICCV= {Int. Conf. Comput. Vis.})

@String(ECCV= {Eur. Conf. Comput. Vis.})

@String(ICLR = {Int. Conf. Learn. Represent.})

@String(PR   = {Pattern Recognition})

@String(CVPRW= {IEEE Conf. Comput. Vis. Pattern Recog. Worksh.})

@String(CVPR  = {CVPR})

@String(ICCV  = {ICCV})

@String(ECCV  = {ECCV})

@String(ICLR  = {ICLR})

@String(PR = {PR})

@String(CVPRW= {CVPRW})

@inproceedings{
carion2026sam3,
title={{SAM} 3: Segment Anything with Concepts},
author={Nicolas Carion and Laura Gustafson and Yuan-Ting Hu and Shoubhik Debnath and Ronghang Hu and Didac Suris Coll-Vinent and Chaitanya Ryali and Kalyan Vasudev Alwala and Haitham Khedr and Andrew Huang and Jie Lei and Tengyu Ma and Baishan Guo and Arpit Kalla and Markus Marks and Joseph Greer and Meng Wang and Peize Sun and Roman R{\"a}dle and Triantafyllos Afouras and Effrosyni Mavroudi and Katherine Xu and Tsung-Han Wu and Yu Zhou and Liliane Momeni and Rishi Hazra and Shuangrui Ding and Sagar Vaze and Francois Porcher and Feng Li and Siyuan Li and Aishwarya Kamath and Ho Kei Cheng and Piotr Dollar and Nikhila Ravi and Kate Saenko and Pengchuan Zhang and Christoph Feichtenhofer},
booktitle={International Conference on Learning Representations (ICLR)},
year={2026},
url={https://openreview.net/forum?id=r35clVtGzw}
}

@inproceedings{chen2025omnire,
  title = {{OmniRe}: Omni Urban Scene Reconstruction},
  author = {Ziyu Chen and Jiawei Yang and Jiahui Huang and Riccardo de Lutio and Janick Martinez Esturo and Boris Ivanovic and Or Litany and Zan Gojcic and Sanja Fidler and Marco Pavone and Li Song and Yue Wang},
  booktitle = {International Conference on Learning Representations (ICLR)},
  year = {2025},
}

@article{chen2025snerfpp,
  title = {{S-NeRF++}: Autonomous Driving Simulation via Neural Reconstruction and Generation},
  author = {Chen, Yurui and Zhang, Junge and Xie, Ziyang and Li, Wenye and Zhang, Feihu and Lu, Jiachen and Zhang, Li},
  journal = {IEEE Transactions on Pattern Analysis and Machine Intelligence},
  year = {2025},
  volume = {47},
  number = {6},
  pages = {4358--4376},
  doi = {10.1109/TPAMI.2025.3543072},
  eprint = {2402.02112},
  archiveprefix = {arXiv},
  primaryclass = {cs.CV},
}

@article{chen2026periodic,
  title = {{Periodic Vibration Gaussian}: Dynamic urban scene reconstruction and real-time rendering},
  author = {Chen, Yurui and Gu, Chun and Jiang, Junzhe and Zhu, Xiatian and Zhang, Li},
  journal = {International Journal of Computer Vision},
  year = {2026},
  volume = {134},
  number = {3},
  pages = {83},
  doi = {10.1007/s11263-026-02740-3},
  eprint = {2311.18561},
  archiveprefix = {arXiv},
  primaryclass = {cs.CV},
}

@article{ferguson2025mhr,
  title={{MHR: Momentum Human Rig}},
  author={Aaron Ferguson and Ahmed A. A. Osman and Berta Besc{\'o}s and Carsten Stoll and Christopher D. Twigg and Christoph Lassner and Dave Otte and {\'E}ric Vignola and Fabi{\'a}n Prada and Federica Bogo and Igor Santesteban and Javier Romero and J. A. Hawing Zarate and Jeongseok Lee and Jinhyung D. Park and Jinlong Yang and John Doublestein and Kishore Venkateshan and Kris Kitani and Ladislav Kavan and Marco Dal Farra and Matthew Hu and Matthew Cioffi and Michele Fabris and MC Ranieri and Mohammad Modarres and Petr Kadlecek and Rawal Khirodkar and Rinat Abdrashitov and Romain Pr{\'e}vost and Rumi Rajbhandari and Ronald Mallet and Russell Pearsall and Sandy Kao and Sanjeev Kumar and Scott Parrish and Shoou-I Yu and Shunsuke Saito and Takaaki Shiratori and Te-Li Wang and Tony Tung and Yi-Chen Xu and Yuan Dong and Yu-Hua Chen and Yuan-Lu Xu and Yuting Ye and Zhongshi Jiang},
  journal={ArXiv},
  year={2025},
  volume={abs/2511.15586},
  url={https://api.semanticscholar.org/CorpusID:283103192}
}

@inproceedings{goel20234dhumans,
  title = {Humans in {4D}: Reconstructing and Tracking Humans with Transformers},
  author = {Goel, Shubham and Pavlakos, Georgios and Rajasegaran, Jathushan and Kanazawa, Angjoo and Malik, Jitendra},
  booktitle = {IEEE/CVF International Conference on Computer Vision (ICCV)},
  year = {2023},
  pages = {14783--14794},
  month = {October},
  eprint = {2305.20091},
  archiveprefix = {arXiv},
  primaryclass = {cs.CV},
}

@inproceedings{hess2025splatad,
  title = {{SplatAD: Real-time lidar and camera rendering with 3D Gaussian splatting for autonomous driving}},
  author = {Hess, Georg and Lindstr{\"o}m, Carl and Fatemi, Maryam and Petersson, Christoffer and Svensson, Lennart},
  booktitle = {Computer Vision and Pattern Recognition Conference (CVPR)},
  year = {2025},
  pages = {11982--11992},
}

@inproceedings{hu2024gauhuman,
  title = {{GauHuman}: Articulated {Gaussian} Splatting from Monocular Human Videos},
  author = {Hu, Shoukang and Hu, Tao and Liu, Ziwei},
  booktitle = {IEEE/CVF Conference on Computer Vision and Pattern Recognition (CVPR)},
  year = {2024},
  pages = {20418--20431},
  month = {June},
  eprint = {2312.02973},
  archiveprefix = {arXiv},
  primaryclass = {cs.CV},
}

@inproceedings{huang2024s3gaussian,
  title = {{$\textit{S}^3$Gaussian}: Self-Supervised Street {Gaussians} for Autonomous Driving},
  author = {Nan Huang and Xiaobao Wei and Wenzhao Zheng and Pengju An and Ming Lu and Wei Zhan and Masayoshi Tomizuka and Kurt Keutzer and Shanghang Zhang},
  booktitle = {IEEE International Conference on Robotics and Automation (ICRA)},
  year = {2026},
}

@article{kerbl20233dgs,
  title = {{3D} {Gaussian} Splatting for Real-Time Radiance Field Rendering},
  author = {Kerbl, Bernhard and Kopanas, Georgios and Leimk{\"u}hler, Thomas and Drettakis, George},
  journal = {ACM Transactions on Graphics},
  year = {2023},
  volume = {42},
  number = {4},
  month = {July},
  url = {https://repo-sam.inria.fr/fungraph/3d-gaussian-splatting/},
}

@inproceedings{kheradmand2024mcmc,
  title = {{3D} {Gaussian} Splatting as Markov Chain Monte Carlo},
  author = {Shakiba Kheradmand and Daniel Rebain and Gopal Sharma and Weiwei Sun and Yang-Che Tseng and Hossam Isack and Abhishek Kar and Andrea Tagliasacchi and Kwang Moo Yi},
  booktitle = {Advances in Neural Information Processing Systems (NeurIPS)},
  year = {2024},
  url = {https://openreview.net/forum?id=UCSt4gk6iX},
}

@inproceedings{kingma2014adam,
  title = {Adam: A method for stochastic optimization},
  author = {Kingma, Diederik P},
  booktitle = {International Conference on Learning Representations (ICLR)},
  year = {2015},
}

@article{krizhevsky2012alexnet,
  title = {{ImageNet} classification with deep convolutional neural networks},
  author = {Krizhevsky, Alex and Sutskever, Ilya and Hinton, Geoffrey E},
  journal = {Advances in Neural Information Processing Systems},
  year = {2012},
  volume = {25},
}

@inproceedings{lei2024gart,
  title = {{GART}: {Gaussian} Articulated Template Models},
  author = {Lei, Jiahui and Wang, Yufu and Pavlakos, Georgios and Liu, Lingjie and Daniilidis, Kostas},
  booktitle = {IEEE/CVF Conference on Computer Vision and Pattern Recognition (CVPR)},
  year = {2024},
  pages = {19876--19887},
  month = {June},
  eprint = {2311.16099},
  archiveprefix = {arXiv},
  primaryclass = {cs.CV},
}

@article{liao2021kitti360,
  title = {{KITTI-360}: A Novel Dataset and Benchmarks for Urban Scene Understanding in {2D} and {3D}},
  author = {Yiyi Liao and Jun Xie and Andreas Geiger},
  journal = {IEEE Transactions on Pattern Analysis and Machine Intelligence},
  year = {2023},
  volume = {45},
  pages = {3292-3310},
  url = {https://api.semanticscholar.org/CorpusID:238198653},
}

@inproceedings{lindstrom2024real2sim,
  title = {Are {NeRFs} ready for autonomous driving? {Towards} closing the real-to-simulation gap},
  author = {Lindstr{\"o}m, Carl and Hess, Georg and Lilja, Adam and Fatemi, Maryam and Hammarstrand, Lars and Petersson, Christoffer and Svensson, Lennart},
  booktitle = {IEEE/CVF Conference on Computer Vision and Pattern Recognition Workshops (CVPRW)},
  year = {2024},
  pages = {4461--4471},
}

@inproceedings{lindstrom2026idsplat,
  title = {{IDSplat}: Instance-Decomposed {3D} {Gaussian} Splatting for Driving Scenes},
  author = {Lindstr{\"o}m, Carl and Rafidashti, Mahan and Fatemi, Maryam and Hammarstrand, Lars and Oswald, Martin R. and Svensson, Lennart},
  booktitle = {IEEE/CVF Conference on Computer Vision and Pattern Recognition Findings (CVPR Findings)},
  year = {2026},
  pages = {316--326},
  month = {June},
  eprint = {2511.19235},
  archiveprefix = {arXiv},
  primaryclass = {cs.CV},
}

@article{loper2015smpl,
  title = {{SMPL}: A Skinned Multi-Person Linear Model},
  author = {Loper, Matthew and Mahmood, Naureen and Romero, Javier and Pons-Moll, Gerard and Black, Michael J.},
  journal = {ACM Transactions on Graphics},
  year = {2015},
  volume = {34},
  number = {6},
  pages = {248:1--248:16},
  doi = {10.1145/2816795.2818013},
}

@inproceedings{ma2025beziergs,
  title = {{BezierGS}: Dynamic Urban Scene Reconstruction with {Bezier} Curve {Gaussian} Splatting},
  author = {Ma, Zipei and Jiang, Junzhe and Chen, Yurui and Zhang, Li},
  booktitle = {IEEE/CVF International Conference on Computer Vision (ICCV)},
  year = {2025},
  pages = {25519-25528},
  eprint = {2506.22099},
  archiveprefix = {arXiv},
}

@inproceedings{mao2026unire,
  title = {{UnIRe}: Unsupervised Instance Decomposition for Dynamic Urban Scene Reconstruction},
  author = {Yunxuan Mao and Rong Xiong and Yue Wang and Yiyi Liao},
  booktitle = {IEEE International Conference on Robotics and Automation (ICRA)},
  year = {2026},
}

@inproceedings{mildenhall2020nerf,
 title={{NeRF}: Representing Scenes as Neural Radiance Fields for View Synthesis},
 author={Ben Mildenhall and Pratul P. Srinivasan and Matthew Tancik and Jonathan T. Barron and Ravi Ramamoorthi and Ren Ng},
 year={2020},
 booktitle={European Conference on Computer Vision (ECCV)},
}

@inproceedings{moon2024exavatar,
  title = {Expressive Whole-Body {3D} {Gaussian} Avatar},
  author = {Moon, Gyeongsik and Shiratori, Takaaki and Saito, Shunsuke},
  booktitle = {European Conference on Computer Vision (ECCV)},
  year = {2024},
  pages = {19--35},
  publisher = {Springer},
  doi = {10.1007/978-3-031-72940-9_2},
  eprint = {2407.21686},
  archiveprefix = {arXiv},
  primaryclass = {cs.CV},
}

@inproceedings{ost2021neuralscenegraphs,
  title = {Neural Scene Graphs for Dynamic Scenes},
  author = {Ost, Julian and Mannan, Fahim and Thuerey, Nils and Knodt, Julian and Heide, Felix},
  booktitle = {IEEE/CVF Conference on Computer Vision and Pattern Recognition (CVPR)},
  year = {2021},
  pages = {2856-2865},
  month = {June},
}

@inproceedings{peng2025desire,
  title = {{DeSiRe-GS: 4D street {Gaussians} for static-dynamic decomposition and surface reconstruction for urban driving scenes}},
  author = {Peng, Chensheng and Zhang, Chengwei and Wang, Yixiao and Xu, Chenfeng and Xie, Yichen and Zheng, Wenzhao and Keutzer, Kurt and Tomizuka, Masayoshi and Zhan, Wei},
  booktitle = {Computer Vision and Pattern Recognition Conference (CVPR)},
  year = {2025},
  pages = {6782--6791},
}

@inproceedings{qian20243dgsavatar,
  title = {{3DGS-Avatar}: Animatable Avatars via Deformable {3D} {Gaussian} Splatting},
  author = {Qian, Zhiyin and Wang, Shaofei and Mihajlovic, Marko and Geiger, Andreas and Tang, Siyu},
  booktitle = {IEEE/CVF Conference on Computer Vision and Pattern Recognition (CVPR)},
  year = {2024},
  pages = {5020--5030},
  month = {June},
  eprint = {2312.09228},
  archiveprefix = {arXiv},
  primaryclass = {cs.CV},
}

@misc{ren2024groundedsam2,
  title = {{Grounded-SAM-2}},
  author = {Ren, Tianhe and Shen, Shuo},
  journal = {GitHub repository},
  year = {2024},
  publisher = {GitHub},
  howpublished = {\url{https://github.com/IDEA-Research/Grounded-SAM-2}},
  commit = {e288b5e72f259e9e26fa5b5cec0df55c4398edc1},
}

@inproceedings{shao2024splattingavatar,
  title = {{SplattingAvatar}: Realistic Real-Time Human Avatars with Mesh-Embedded Gaussian Splatting},
  author = {Shao, Zhijing and Wang, Zhaolong and Li, Zhuang and Wang, Duotun and Lin, Xiangru and Zhang, Yu and Fan, Mingming and Wang, Zeyu},
  booktitle = {IEEE/CVF Conference on Computer Vision and Pattern Recognition (CVPR)},
  year = {2024},
  eprint = {2403.05087},
  archiveprefix = {arXiv},
}

@article{simeoni2025dinov3,
  title={Dinov3},
  author={Sim{\'e}oni, Oriane and Vo, Huy V and Seitzer, Maximilian and Baldassarre, Federico and Oquab, Maxime and Jose, Cijo and Khalidov, Vasil and Szafraniec, Marc and Yi, Seungeun and Ramamonjisoa, Micha{\"e}l and others},
  journal={arXiv preprint arXiv:2508.10104},
  year={2025}
}

@inproceedings{song2025coda,
  title = {{CoDa-4DGS}: Dynamic {Gaussian} splatting with context and deformation awareness for autonomous driving},
  author = {Song, Rui and Liang, Chenwei and Xia, Yan and Zimmer, Walter and Cao, Hu and Caesar, Holger and Festag, Andreas and Knoll, Alois},
  booktitle = {IEEE/CVF International Conference on Computer Vision (ICCV)},
  year = {2025},
  publisher = {IEEE/CVF},
}

@inproceedings{su2025dial,
  title = {{DIAL-GS}: Dynamic Instance Aware Reconstruction for Label-free Street Scenes with 4D Gaussian Splatting},
  author = {Su, Chenpeng and Wu, Wenhua and Peng, Chensheng and Deng, Tianchen and Liu, Zhe and Wang, Hesheng},
  booktitle = {IEEE International Conference on Robotics and Automation (ICRA)},
  year = {2026},
  doi = {10.1109/icra57385.2026.11696761},
}

@inproceedings{sun2020waymo,
  title = {Scalability in Perception for Autonomous Driving: Waymo Open Dataset},
  author = {Sun, Pei and Kretzschmar, Henrik and Dotiwalla, Xerxes and Chouard, Aurelien and Patnaik, Vijaysai and Tsui, Paul and Guo, James and Zhou, Yin and Chai, Yuning and Caine, Benjamin and Vasudevan, Vijay and Han, Wei and Ngiam, Jiquan and Zhao, Hang and Timofeev, Aleksei and Ettinger, Scott and Krivokon, Maxim and Gao, Amy and Joshi, Aditya and Zhang, Yu and Shlens, Jonathon and Chen, Zhifeng and Anguelov, Dragomir},
  booktitle = {IEEE/CVF Conference on Computer Vision and Pattern Recognition (CVPR)},
  year = {2020},
  month = {June},
}

@inproceedings{sun2025splatflow,
  title = {{SplatFlow}: Self-Supervised Dynamic Gaussian Splatting in Neural Motion Flow Field for Autonomous Driving},
  author = {Sun, Su and Zhao, Cheng and Sun, Zhuoyang and Chen, Yingjie Victor and Chen, Mei},
  booktitle = {IEEE/CVF Conference on Computer Vision and Pattern Recognition (CVPR)},
  year = {2025},
  pages = {27487-27496},
  month = {June},
}

@inproceedings{tancik2022blocknerf,
  title = {{Block-NeRF}: Scalable Large Scene Neural View Synthesis},
  author = {Tancik, Matthew and Casser, Vincent and Yan, Xinchen and Pradhan, Sabeek and Mildenhall, Ben and Srinivasan, Pratul P. and Barron, Jonathan T. and Kretzschmar, Henrik},
  booktitle = {IEEE/CVF Conference on Computer Vision and Pattern Recognition (CVPR)},
  year = {2022},
  pages = {8248-8258},
  month = {June},
}

@inproceedings{tonderski2024neurad,
  title = {{NeuRAD: Neural Rendering for Autonomous Driving}},
  author = {Tonderski, Adam and Lindstr{\"o}m, Carl and Hess, Georg and Ljungbergh, William and Svensson, Lennart and Petersson, Christoffer},
  booktitle = {IEEE/CVF Conference on Computer Vision and Pattern Recognition (CVPR)},
  year = {2024},
  pages = {14895-14904},
  publisher = {IEEE Computer Society},
  address = {Los Alamitos, CA, USA},
  month = {Jun},
  doi = {10.1109/CVPR52733.2024.01411},
  url = {https://doi.ieeecomputersociety.org/10.1109/CVPR52733.2024.01411},
}

@inproceedings{tourani2025sdf,
  title = {Leveraging {2D} Priors and {SDF} Guidance for Dynamic Urban Scene Rendering},
  author = {Tourani, Siddharth and Reddy, Jayaram and Kumbar, Akash and Tourani, Satyajit and Goyal, Nishant and Krishna, Madhava and Reddy, N. Dinesh and Khan, Muhammad Haris},
  booktitle = {IEEE/CVF International Conference on Computer Vision (ICCV)},
  year = {2025},
  eprint = {2510.13381},
  archiveprefix = {arXiv},
}

@InProceedings{turki2023suds,
                    title = {{SUDS}: Scalable Urban Dynamic Scenes},
                    author = {Turki, Haithem and Zhang, Jason Y and Ferroni, Francesco and Ramanan, Deva},
                    booktitle = {Computer Vision and Pattern Recognition (CVPR)},
                    year = {2023}
                }

@article{wang2004ssim,
  title = {Image quality assessment: from error visibility to structural similarity},
  author = {Zhou Wang and Bovik, A.C. and Sheikh, H.R. and Simoncelli, E.P.},
  journal = {IEEE Transactions on Image Processing},
  year = {2004},
  volume = {13},
  number = {4},
  pages = {600-612},
  doi = {10.1109/TIP.2003.819861},
}

@inproceedings{wei2025emd,
  title = {{EMD}: Explicit Motion Modeling for High-Quality Street Gaussian Splatting},
  author = {Wei, Xiaobao and Wuwu, Qingpo and Zhao, Zhongyu and Wu, Zhuangzhe and Huang, Nan and Lu, Ming and Ma, Ningning and Zhang, Shanghang},
  booktitle = {IEEE/CVF International Conference on Computer Vision (ICCV)},
  year = {2025},
  pages = {28462-28472},
  eprint = {2411.15582},
  archiveprefix = {arXiv},
}

@inproceedings{wu20244dgs,
  title = {{4D} {Gaussian} Splatting for Real-Time Dynamic Scene Rendering},
  author = {Wu, Guanjun and Yi, Taoran and Fang, Jiemin and Xie, Lingxi and Zhang, Xiaopeng and Wei, Wei and Liu, Wenyu and Tian, Qi and Wang, Xinggang},
  booktitle = {IEEE/CVF Conference on Computer Vision and Pattern Recognition (CVPR)},
  year = {2024},
  pages = {20310-20320},
  month = {June},
}

@inproceedings{wu2024mars,
  title = {{MARS}: {An} Instance-Aware, Modular and Realistic Simulator for Autonomous Driving},
  author = {Wu, Zirui
and Liu, Tianyu
and Luo, Liyi
and Zhong, Zhide
and Chen, Jianteng
and Xiao, Hongmin
and Hou, Chao
and Lou, Haozhe
and Chen, Yuantao
and Yang, Runyi
and Huang, Yuxin
and Ye, Xiaoyu
and Yan, Zike
and Shi, Yongliang
and Liao, Yiyi
and Zhao, Hao},
  editor = {Fang, Lu
and Pei, Jian
and Zhai, Guangtao
and Wang, Ruiping},
  booktitle = {Artificial Intelligence},
  year = {2024},
  pages = {3--15},
  publisher = {Springer Nature Singapore},
  address = {Singapore},
  isbn = {978-981-99-8850-1},
}

@inproceedings{xiao2021pandaset,
  title = {{PandaSet}: Advanced Sensor Suite Dataset for Autonomous Driving},
  author = {Xiao, Pengchuan and Shao, Zhenlei and Hao, Steven and Zhang, Zishuo and Chai, Xiaolin and Jiao, Judy and Li, Zesong and Wu, Jian and Sun, Kai and Jiang, Kun and Wang, Yunlong and Yang, Diange},
  booktitle = {IEEE International Intelligent Transportation Systems Conference (ITSC)},
  year = {2021},
  pages = {3095-3101},
  doi = {10.1109/ITSC48978.2021.9565009},
}

@inproceedings{xie2023snerf,
  title = {S-{N}e{RF}: Neural Radiance Fields for Street Views},
  author = {Ziyang Xie and Junge Zhang and Wenye Li and Feihu Zhang and Li Zhang},
  booktitle = {International Conference on Learning Representations (ICLR)},
  year = {2023},
  url = {https://openreview.net/forum?id=gx2yJS-ENqI},
}

@inproceedings{xu2025ad,
  title = {{AD-GS}: Object-Aware {B-Spline} {Gaussian} Splatting for Self-Supervised Autonomous Driving},
  author = {Xu, Jiawei and Deng, Kai and Fan, Zexin and Wang, Shenlong and Xie, Jin and Yang, Jian},
  booktitle = {International Conference on Computer Vision (ICCV)},
  year = {2025},
}

@inproceedings{yan2024streetgs,
  title = {Street {Gaussians} for Modeling Dynamic Urban Scenes},
  author = {Yunzhi Yan and Haotong Lin and Chenxu Zhou and Weijie Wang and Haiyang Sun and Kun Zhan and Xianpeng Lang and Xiaowei Zhou and Sida Peng},
  booktitle = {European Conference on Computer Vision (ECCV)},
  year = {2024},
}

@inproceedings{yang2023emernerf,
  title = {{EmerNeRF}: Emergent spatial-temporal scene decomposition via self-supervision},
  author = {Yang, Jiawei and Ivanovic, Boris and Litany, Or and Weng, Xinshuo and Kim, Seung Wook and Li, Boyi and Che, Tong and Xu, Danfei and Fidler, Sanja and Pavone, Marco and others},
  booktitle = {International Conference on Learning Representations (ICLR)},
  year = {2024},
}

@inproceedings{yang2026sam3dbody,
  title = {{SAM} {3D} Body: Robust Full-Body Human Mesh Recovery},
  author = {Yang, Xitong and Kukreja, Devansh and Pinkus, Don and Fan, Taosha and Park, Jinhyung and Shin, Soyong and Cao, Jinkun and Liu, Jia-Wei and Ugrinovic, Nicol{\'a}s and Sagar, Anushka and Malik, Jitendra and Feiszli, Matt and Doll{\'a}r, Piotr and Kitani, Kris},
  booktitle = {IEEE/CVF Conference on Computer Vision and Pattern Recognition (CVPR)},
  year = {2026},
  pages = {7209--7219},
  month = {June},
  eprint = {2602.15989},
  archiveprefix = {arXiv},
  primaryclass = {cs.CV},
}

@inproceedings{zhang2018lpips,
  title = {The Unreasonable Effectiveness of Deep Features as a Perceptual Metric},
  author = {Zhang, Richard and Isola, Phillip and Efros, Alexei A. and Shechtman, Eli and Wang, Oliver},
  booktitle = {IEEE Conference on Computer Vision and Pattern Recognition (CVPR)},
  year = {2018},
  month = {June},
}

@inproceedings{zhou2024drivinggaussian,
  title = {{DrivingGaussian}: Composite {Gaussian} splatting for surrounding dynamic autonomous driving scenes},
  author = {Zhou, Xiaoyu and Lin, Zhiwei and Shan, Xiaojun and Wang, Yongtao and Sun, Deqing and Yang, Ming-Hsuan},
  booktitle = {IEEE/CVF Conference on Computer Vision and Pattern Recognition (CVPR)},
  year = {2024},
  pages = {21634--21643},
}

@inproceedings{zhou2024hugs,
  title = {{HUGS}: Holistic Urban 3D Scene Understanding via Gaussian Splatting},
  author = {Zhou, Hongyu and Shao, Jiahao and Xu, Lu and Bai, Dongfeng and Qiu, Weichao and Liu, Bingbing and Wang, Yue and Geiger, Andreas and Liao, Yiyi},
  booktitle = {IEEE/CVF Conference on Computer Vision and Pattern Recognition (CVPR)},
  year = {2024},
  pages = {21336-21345},
  month = {June},
}

@inproceedings{osman2020star,
      author = {Osman, Ahmed A A and Bolkart, Timo and Black, Michael J.},
      title = {{STAR}: A Sparse Trained Articulated Human Body Regressor},
      booktitle = {European Conference on Computer Vision (ECCV)},
      pages = {598--613},
      year = {2020},
      url = {https://star.is.tue.mpg.de}
}

@inproceedings{xu2020ghumghuml,
  author={Xu, Hongyi and Bazavan, Eduard Gabriel and Zanfir, Andrei and Freeman, William T. and Sukthankar, Rahul and Sminchisescu, Cristian},
  booktitle={IEEE/CVF Conference on Computer Vision and Pattern Recognition (CVPR)}, 
  title={{GHUM \& GHUML}: Generative 3D Human Shape and Articulated Pose Models}, 
  year={2020},
  volume={},
  number={},
  pages={6183-6192},
  doi={10.1109/CVPR42600.2020.00622}
}

@article{guo2023streetsurf,
  title={{StreetSurf}: Extending Multi-view Implicit Surface Reconstruction to Street Views},
  author={Jianfei Guo and Nianchen Deng and Xinyang Li and Ye-Qi Bai and Botian Shi and Chi-Yu Wang and Chenjing Ding and Dongliang Wang and Yi-Kang Li},
  journal={ArXiv},
  year={2023},
  volume={abs/2306.04988},
  url={https://api.semanticscholar.org/CorpusID:259108796}
}

@inproceedings{geiger2012kitti,
  title = {Are we ready for autonomous driving? {The} {KITTI} vision benchmark suite},
  author = {Geiger, Andreas and Lenz, Philip and Urtasun, Raquel},
  booktitle = {IEEE Conference on Computer Vision and Pattern Recognition (CVPR)},
  pages = {3354--3361},
  year = {2012},
  doi = {10.1109/CVPR.2012.6248074},
}

@inproceedings{rempe2021humor,
  title = {{HuMoR}: {3D} Human Motion Model for Robust Pose Estimation},
  author = {Rempe, Davis and Birdal, Tolga and Hertzmann, Aaron and Yang, Jimei and Sridhar, Srinath and Guibas, Leonidas J.},
  booktitle = {IEEE/CVF International Conference on Computer Vision (ICCV)},
  year = {2021},
  pages = {11468--11479},
  doi = {10.1109/ICCV48922.2021.01129},
}

@article{besl1992icp,
  title = {A Method for Registration of 3-{D} Shapes},
  author = {Besl, Paul J. and McKay, Neil D.},
  journal = {IEEE Transactions on Pattern Analysis and Machine Intelligence},
  year = {1992},
  volume = {14},
  number = {2},
  pages = {239--256},
  doi = {10.1109/34.121791},
}

@article{harvey2020inbetweening,
  title = {Robust Motion In-betweening},
  author = {Harvey, F{\'e}lix G. and Yurick, Mike and Nowrouzezahrai, Derek and Pal, Christopher},
  journal = {ACM Transactions on Graphics},
  year = {2020},
  volume = {39},
  number = {4},
  doi = {10.1145/3386569.3392480},
}

@inproceedings{tevet2023mdm,
  title = {Human Motion Diffusion Model},
  author = {Tevet, Guy and Raab, Sigal and Gordon, Brian and Shafir, Yoni and Cohen-Or, Daniel and Bermano, Amit H.},
  booktitle = {International Conference on Learning Representations (ICLR)},
  year = {2023},
}

@inproceedings{cohan2024condmdi,
  title = {Flexible Motion In-betweening with Diffusion Models},
  author = {Cohan, Setareh and Tevet, Guy and Reda, Daniele and Peng, Xue Bin and van de Panne, Michiel},
  booktitle = {ACM SIGGRAPH Conference Papers},
  year = {2024},
  pages = {1--9},
  doi = {10.1145/3641519.3657414},
}

@article{everingham2010pascal,
  title = {The {PASCAL} Visual Object Classes ({VOC}) Challenge},
  author = {Everingham, Mark and Van Gool, Luc and Williams, Christopher K. I. and Winn, John and Zisserman, Andrew},
  journal = {International Journal of Computer Vision},
  volume = {88},
  number = {2},
  pages = {303--338},
  year = {2010},
}

@inproceedings{lin2014coco,
  title = {Microsoft {COCO}: Common Objects in Context},
  author = {Lin, Tsung-Yi and Maire, Michael and Belongie, Serge and Hays, James and Perona, Pietro and Ramanan, Deva and Doll{\'a}r, Piotr and Zitnick, C. Lawrence},
  booktitle = {European Conference on Computer Vision (ECCV)},
  pages = {740--755},
  year = {2014},
}

@article{yang2013articulated,
  title = {Articulated Human Detection with Flexible Mixtures of Parts},
  author = {Yang, Yi and Ramanan, Deva},
  journal = {IEEE Transactions on Pattern Analysis and Machine Intelligence},
  volume = {35},
  number = {12},
  pages = {2878--2890},
  year = {2013},
}

@inproceedings{li2022vitdet,
  title = {Exploring Plain Vision Transformer Backbones for Object Detection},
  author = {Li, Yanghao and Mao, Hanzi and Girshick, Ross and He, Kaiming},
  booktitle = {European Conference on Computer Vision (ECCV)},
  year = {2022},
}

@inproceedings{xu2022vitpose,
  title = {{ViTPose}: Simple Vision Transformer Baselines for Human Pose Estimation},
  author = {Xu, Yufei and Zhang, Jing and Zhang, Qiming and Tao, Dacheng},
  booktitle = {Advances in Neural Information Processing Systems (NeurIPS)},
  year = {2022},
}

@inproceedings{liu2024groundingdino,
  title = {Grounding {DINO}: Marrying {DINO} with Grounded Pre-Training for Open-Set Object Detection},
  author = {Liu, Shilong and Zeng, Zhaoyang and Ren, Tianhe and Li, Feng and Zhang, Hao and Yang, Jie and Jiang, Qing and Li, Chunyuan and Yang, Jianwei and Su, Hang and Zhu, Jun and Zhang, Lei},
  booktitle = {European Conference on Computer Vision (ECCV)},
  year = {2024},
}

@ARTICLE{zhou2026hugsim,
  author={Zhou, Hongyu and Lin, Longzhong and Wang, Jiabao and Lu, Yichong and Bai, Dongfeng and Liu, Bingbing and Wang, Yue and Geiger, Andreas and Liao, Yiyi},
  journal={IEEE Transactions on Pattern Analysis and Machine Intelligence}, 
  title={{HUGSIM}: A Real-Time, Photo-Realistic and Closed-Loop Simulator for Autonomous Driving}, 
  year={2026},
  volume={48},
  number={4},
  pages={4673-4691},
  doi={10.1109/TPAMI.2025.3647952}}

@inproceedings{lewis2000psd,
  title = {Pose Space Deformation: A Unified Approach to Shape Interpolation and Skeleton-Driven Deformation},
  author = {Lewis, J. P. and Cordner, Matt and Fong, Nickson},
  booktitle = {Annual Conference on Computer Graphics and Interactive Techniques (SIGGRAPH)},
  year = {2000},
  pages = {165--172},
}

@inproceedings{kavan2007dqs,
  title = {Skinning with Dual Quaternions},
  author = {Kavan, Ladislav and Collins, Steven and {\v{Z}}{\'a}ra, Ji{\v{r}}{\'\i} and O'Sullivan, Carol},
  booktitle = {Symposium on Interactive 3D Graphics and Games (I3D)},
  year = {2007},
  pages = {39--46},
}

@inproceedings{shoemake1985slerp,
  title = {Animating Rotation with Quaternion Curves},
  author = {Shoemake, Ken},
  booktitle = {Annual Conference on Computer Graphics and Interactive Techniques (SIGGRAPH)},
  year = {1985},
  pages = {245--254},
}

@inproceedings{magnenatthalmann1988lbs,
  title = {Joint-Dependent Local Deformations for Hand Animation and Object Grasping},
  author = {Magnenat-Thalmann, Nadia and Laperri{\`e}re, Richard and Thalmann, Daniel},
  booktitle = {Graphics Interface (GI)},
  year = {1988},
  pages = {26--33},
}

@article{oreshkin2024deltainterpolator,
  title = {Motion In-Betweening via Deep $\Delta$-Interpolator},
  author = {Oreshkin, Boris N. and Valkanas, Antonios and Harvey, F{\'e}lix G. and M{\'e}nard, Louis-Simon and Bocquelet, Florent and Coates, Mark J.},
  journal = {IEEE Transactions on Visualization and Computer Graphics},
  year = {2024},
  volume = {30},
  number = {8},
  pages = {5693--5704},
  doi = {10.1109/TVCG.2023.3309107},
}
}

\clearpage
\setcounter{page}{1}
\maketitlesupplementary
\appendix

\newcounter{suppref}
\renewcommand{\thesuppref}{S\arabic{suppref}}
\newcommand{\citeS}[1]{[\ref{suppref:#1}]}
\newenvironment{supprefs}{\begin{list}{[\thesuppref]}{\usecounter{suppref}\setlength{\labelwidth}{2.2em}%
  \setlength{\leftmargin}{2.8em}\setlength{\itemsep}{2pt}\small}}{\end{list}}

\noindent In this supplementary material, we provide further details of our implementation and evaluation, as well as additional quantitative and qualitative results. \Cref{sec:supmat_impl} describes the implementation, including the refinement of the instance masks, the selection of pedestrians, the body frame, the initialization of the Gaussians, the interpolation network and the training settings. \Cref{sec:supmat_dataeval} describes the datasets, the evaluation settings and metrics, and the sources of the baseline results. \Cref{sec:supmat_results} reports results on dynamic NOTR, the SplatFlow scenes and KITTI, and \cref{sec:supmat_qualitative} shows further qualitative results on WOD. \Cref{sec:supmat_interp} studies the learned interpolation with fewer training frames, including a comparison with CondMDI. \Cref{sec:supmat_downstream} gives the details of the downstream perception evaluation, and \cref{sec:supmat_reposing} those of the re-posing. \Cref{sec:supmat_runtime} reports preprocessing, optimization and inference times, with a breakdown over the components of our method. Finally, \cref{sec:supmat_limitations} discusses the limitations of our method in more detail.

\section{Implementation details}
\label{sec:supmat_impl}

\parsection{Instance masks}
We run SAM~3~\cite{carion2026sam3} on the frames of each camera separately, with one detection and tracking pass per text prompt: \emph{car}, \emph{truck}, \emph{bus}, \emph{train}, \emph{human}, \emph{van}, \emph{cyclist}, \emph{bicycle} and \emph{pedestrian}, the categories IDSplat uses with Grounded-SAM~2~\cite{ren2024groundedsam2}. A new object is added as a track if its detection score is at least 0.5 (instead of the default 0.7), and during the hot-start period of 15 frames, a new track is removed only when it is not detected for 12 frames instead of 8. Both changes make it easier to track distant objects. The masklets of a camera are refined in three steps and then merged across cameras.

\emph{Drift splitting:} 
A masklet is separated into two where it reappears after at least four frames without a mask, with a box $\mathbf{b}$, inconsistent with its last box $\mathbf{b}'$. This is when $\mathrm{IoU}(\mathbf{b}, \mathbf{b}') < 0.1$, and either the box area changes by more than a factor of four or the box center moves by more than $1.5\sqrt{\max(A, A')}$, where $A$ and $A'$ are the box areas. The class of the separated masklets is then marked as unreliable, since it belongs to the object the class started on. This helps in cases where for example a human mask drifts onto a car or some stationary object in the scene.

\emph{Merging:} 
Duplicate masklets are merged in two passes over pairs of masklets that share at least half of the frames of the shorter one (at most 20 shared frames are compared). In the first pass, masklets with an IoU of at least 0.5 are merged into the more confident one. In the second pass, a masklet that lies almost entirely inside a larger one, with an average of at least 80\% of it's pixels covered by the larger masklet over the frames they share, is merged into it, since it is likely of the same object. A frame where only the merged smaller masklet exists are added to the merged track, only if it has overlap with its closest dilated masklets in time, from that track. The masklet takes the class of the most confident of the merged pair.

\emph{Attachment:}
A piece that was split off by drift splitting and was not merged is joined to a reliable masklet if, on the frames they share, on average at least half of its pixels lie within 10 pixels of that masklet. This merges, for example, a backpack that a pedestrian is wearing.

\emph{Cross-camera merging:}
Across cameras, masklets are associated through their lidar points as in IDSplat. Each lidar point of a sweep is projected into every camera's image plane and assigned to the masklet whose mask, eroded by 3 pixels, contains it, so a point in the overlap of two camera views is assigned to a masklet in each. Two masklets that are assigned more than 50 of the same lidar points, counted over all sweeps, are merged as views of the same object. We process these pairs from the most to the fewest shared points and refuse a merge that would join two masklets seen by the same camera at the same time, since SAM~3 gives different objects in one image different identities. Each merged instance then takes the class it has in most of its frames, counted once per lidar frame. Instances with fewer than three successfully registered frames are moved to the static background.

\parsection{Pedestrian selection}
\sambody{}~\cite{yang2026sam3dbody} is applied to every masklet of the classes \emph{human}, \emph{pedestrian} and \emph{cyclist}, prompted with the mask and its bounding box, using the camera intrinsics of the dataset and only the body decoder. The mesh is aligned with the lidar points of the instance in that frame by a translation estimated with ICP~\cite{besl1992icp} from the half of the mesh vertices closest to the camera, which are the ones the lidar observes.
The translation is initialized with the difference of the centroids and updated from nearest-neighbor matches until it changes by less than 0.1\,mm, for at most 20 iterations.
We then check whether the mesh is consistent with the lidar points. In each frame where the body is within 80\,m of the camera, we compute the fraction of the lidar points of the instance that lie within 0.2\,m of a mesh vertex. If the median fraction over these frames is at least 0.5, the instance is treated as a pedestrian. Otherwise, it is kept as a rigid object if the median number of lidar points per frame is at least 20 (which retains misclassified objects such as vehicles) and removed if it is lower. When body estimation fails or a pedestrian is occluded, the frame is treated as missing and its pose is interpolated (see \emph{Interpolation network} below). IDSplat assigns instances whose trajectory barely moves to the static background. Pedestrians with a body estimate are kept as instances regardless.

\parsection{Body frame}
\sambody{} predicts the body in centimeters, in a frame whose origin is the top-level joint of \mhr{}~\cite{ferguson2025mhr} and whose axes are those of the observing camera ($x$ right, $y$ up, $z$ towards the camera), together with the position of this origin relative to the camera. The transform from the local frame of the instance to the body frame (\cref{sec:pedestrian_init}) is
\begin{equation}
  \mathbf{M} = \mathbf{S}\,\mathbf{D}^{-1}\,\mathbf{C}_c^{-1}\,\mathbf{P}_c,
  \label{eq:supmat_bodyframe}
\end{equation}
where $\mathbf{P}_c$ is the instance pose and $\mathbf{C}_c$ the camera-to-world pose of the observing camera with the axes above, both at the canonical frame $c$. $\mathbf{D}$ is the translation by the predicted body position, with its $y$ and $z$ components negated to convert from the OpenCV axes of \sambody{}. $\mathbf{S}=\mathrm{diag}(100,100,100,1)$ converts meters to centimeters. $\mathbf{M}$ is a similarity transform with uniform scale, so $\mathbf{M}^{-1}$ maps a posed point back to the local frame. The joint transformations $\mathbf{T}^j_c$ are those of the \sambody{} estimate at $c$ with the median skeleton (\cref{sec:body_est}).

\parsection{Gaussian initialization}
The 9220 body Gaussians are placed at every second vertex of the \mhr{} mesh at the canonical frame (\cref{sec:pedestrian_init}). Their colors are initialized to gray; opacities and scales are initialized as for all other Gaussians.
Lidar Gaussians are initialized from the lidar points $\mathbf{p}$ of the pedestrian at the canonical frame, expressed in its local frame. Each point $\mathbf{p}$ is assigned the vertex $i(\mathbf{p})$ of its nearest body Gaussian,
\begin{equation}
    i(\mathbf{p}) = \operatorname*{arg\,min}_{i} \left\| \mathbf{p} - \boldsymbol{\mu}^i \right\|_2,
\end{equation}
where $\boldsymbol{\mu}^i$ is the initial mean of the body Gaussian at vertex $i$. Points farther than 0.3\,m from the body Gaussian of vertex $i(\mathbf{p})$ are discarded, and at most 2000 points per pedestrian are kept, chosen at random. The Gaussian initialized from $\mathbf{p}$ takes $i(\mathbf{p})$ as its source vertex $i$ in \cref{eq:lbs}, so it uses the skinning weights of that vertex and moves with the body part it lies on. Its color is initialized from the camera image, as for the other lidar points.
During MCMC densification~\citeS{kheradmand2024mcmc}, a Gaussian that is relocated or added copies the instance identity and the vertex index of its source, and thus its skinning weights. As in IDSplat, instances without a body receive 1000 additional Gaussians at random positions inside their bounding box, but pedestrians do not.

\parsection{Pedestrian heading}
\sambody{} estimates the global rotation of the body relative to the observing camera, so the joint transformations $\mathbf{T}^j_t$ already contain changes in the heading of the pedestrian. Applying the rotation of the trajectory $\mathbf{P}_t$ as well would count them twice. The pose $\tilde{\mathbf{P}}_t$ in the body deformation (\cref{sec:motion}) therefore takes its rotation from the camera instead of the trajectory,
\begin{align}
    & \tilde{\mathbf{P}}_t\,\mathbf{x} = \tilde{\mathbf{R}}_t\left(\mathbf{x} - \mathbf{o}\right) + \mathbf{P}_t\,\mathbf{o},\\
    & \tilde{\mathbf{R}}_t = \mathbf{R}(\mathbf{C}_t)\,\mathbf{R}(\mathbf{C}_c)^\top\,\mathbf{R}(\mathbf{P}_c),
\end{align}
where $\mathbf{R}(\cdot)$ is the rotation of a transform and $\mathbf{C}_t$ the pose of the camera that observed the pedestrian at $t$, interpolated with SLERP~\cite{shoemake1985slerp} between frames and held constant outside them, and $\mathbf{o}$ is the canonical position of the pelvis in the local frame. The body thus turns about its pelvis, which follows the trajectory. 
Gaussians also rotate with the motion. A body Gaussian of vertex $i$ is rotated by the rotation closest to $\sum_j w^{ij}\mathbf{R}(\mathbf{T}^j_t)\,\mathbf{R}(\mathbf{T}^j_c)^\top$, mapped to the local frame, followed by $\tilde{\mathbf{R}}_t$. The Gaussians of all other dynamic instances are rotated with their trajectory. In IDSplat, the orientations of dynamic Gaussians stay fixed in the local frame and only their means move.

\parsection{Body refinement}
The \mhr{} model has 204 parameters: 136 pose parameters, consisting of the global translation (3), the global rotation (3), the joint angles (124), and the translational parameters (6), followed by 68 skeleton parameters $\boldsymbol{\sigma}$. The 45 shape coefficients of \sambody{} only change the surface and are fixed to their median values for each track. 
For each pedestrian, we optimize the offset $\Delta\mathbf{z}\in\mathbb{R}^{28}$ to the skeleton coefficients (\cref{sec:motion}) and an offset to the six translational pose parameters, which lengthen the spine, neck, clavicles, arms, hips and legs and are therefore initialized with their median over the track, like the skeleton. For each observed frame $t$, we optimize the offset $\Delta\boldsymbol{\theta}_t\in\mathbb{R}^{127}$ to the global rotation and the joint angles. The global translation is not optimized, since the trajectory places the pedestrian.
The offsets are optimized with Adam~\citeS{kingma2014adam} together with all other parameters (\cref{tab:supmat_lr}).

\parsection{Interpolation network}
In this work, frames in which a pedestrian has a body estimate are called observations, that is, the training frames where its masklet is tracked and \sambody{} predicts a body. Held-out frames have no body estimates, and neither do training frames where the pedestrian is fully occluded, not tracked, or does not have a \sambody{} estimate. 
At a time $t$ without a body estimate (\ie at a held-out frame), let $a$ and $b$ be the observations before and after $t$, and $\alpha\in[0,1]$ the position of $t$ between them. Before the first and after the last observation, the nearest one is used. The joint angles $\boldsymbol{\phi}_t\in\mathbb{R}^{124}$, part of the pose parameters $\boldsymbol{\theta}_t$, are interpolated as follows
\begin{equation}
    \boldsymbol{\phi}_t = \boldsymbol{\phi}_a + \alpha\,\mathrm{wrap}\big(\boldsymbol{\phi}_b - \boldsymbol{\phi}_a\big),
\end{equation}
where $\mathrm{wrap}$ maps angles to $[-\pi, \pi)$. The remaining parameters are interpolated linearly. The global rotation is interpolated with SLERP between the rotations it applies to the body about the pelvis. Since \mhr{} maps its parameters linearly to the Euler angles of the joints, this interpolates every joint angle linearly. 
The correction network adds $\Delta\boldsymbol{\phi}_t = 4\alpha(1-\alpha)\,f(\cdot)$ to the interpolated joint angles, so the correction vanishes at observations. It does not change the global rotation. Its input is a window of four observations $k = 1, \ldots, 4$ in time order, $a$ and $b$ around the gap, and one more before and after them. A shared encoder maps each observation to a 128-dimensional feature vector from the sine and cosine of its joint angles $\sin\boldsymbol{\phi}_k$ and $\cos\boldsymbol{\phi}_k$, its global rotation relative to that of $a$ as a rotation vector, its time offset $t_k-t$, both unprocessed and as 32 pairs $\sin(2\pi\nu_m(t_k - t))$, $\cos(2\pi\nu_m(t_k - t))$ with periods $1/\nu_m$ spaced geometrically from 0.2\,s to 4\,s, and a flag that marks whether it is a separate observation (at the end of a track, the outer observation repeats $a$ or $b$). The encoder has two layers with ReLU activations. An MLP with two hidden layers of 256 units and ReLU activations maps the four features, in time order, together with the joint angular velocities between consecutive observations, $\mathrm{wrap}(\boldsymbol{\phi}_{k+1} - \boldsymbol{\phi}_k)/(t_{k+1}-t_k)$, to $f\in\mathbb{R}^{124}$. Its last layer is initialized to zero (0.38\,M parameters in total). Global rotations of observations from different cameras are first brought to a common frame by the yaw of the camera.
The network is applied to pedestrians whose mean speed over the track is at least 0.5\,m/s, for times more than 0.02\,s from both observations in gaps of at most 0.65\,s, and only if the global rotations of consecutive observations in the window differ by at most 60\textdegree{}. The network is used at held-out frames in NVS evaluation. 

\emph{Training:}
The training data are the \sambody{} estimates of the 39 NOTR~\cite{yang2023emernerf} scenes including pedestrians with body estimates, none of which is one of the eight OmniRe evaluation scenes, processed as in our method: tracks of at least nine frames with the median skeleton per track. Only tracks in which some joint has a standard deviation of at least 12\,cm relative to the pelvis are used, which removes pedestrians whose pose barely changes, leaving 92{,}787 poses. We use the 8 OmniRe scenes and the 7 NOTR scenes not included in NOTR dynamic for the evaluations on the NOTR dynamic data. 
A training window hides one frame $t$ and takes $a$ and $b$ two to six frames apart around it, and the outer observations one to four frames beyond $a$ and $b$, chosen independently, to cover the irregular spacing of observations at render time.
In 10\% of the windows, an outer observation is replaced by $a$ or $b$, as at the end of a track. The loss is the mean distance between the joint positions computed by \mhr{} from the corrected and the hidden pose, relative to the pelvis. We train for 20{,}000 steps with batches of 512 windows, using AdamW~\citeS{loshchilov2019adamw} with a learning rate of $10^{-3}$, cosine decay and a weight decay of $10^{-3}$.

\parsection{Scene optimization} 
Each scene is optimized for 30{,}000 steps with the losses, weights, learning rates, resolution schedule, and MCMC densification of IDSplat. No loss is added for pedestrians. \Cref{tab:supmat_lr} lists the learning rates of the body parameters. 

\begin{table}[t]
  \centering
  \caption{Loss weights, shared with IDSplat~\cite{lindstrom2026idsplat}, which takes the losses from SplatAD~\cite{hess2025splatad}. The MCMC regularization follows~\citeS{kheradmand2024mcmc}.}
  \label{tab:supmat_losses}
  \footnotesize
  \begin{tabular}{@{}lc@{}}
    \toprule
    Loss & Weight \\
    \midrule
    Image L1 & $0.8$ \\
    Image D-SSIM & 0.2 \\
    Lidar depth & 0.1 \\
    Lidar intensity & 1.0 \\
    Lidar ray drop & 0.1 \\
    Lidar line of sight & 0.1 \\
    MCMC opacity regularization & 0.005 \\
    MCMC scale regularization & 0.001 \\
    \bottomrule
  \end{tabular}
\end{table}

\begin{table}[t]
  \centering
  \caption{Learning rates of the body parameters, in addition to those of IDSplat~\cite{lindstrom2026idsplat}. Adam~\citeS{kingma2014adam}; after a warm-up from $10^{-8}$ over the given number of steps, the learning rate decays exponentially to its final value at the last step.}
  \label{tab:supmat_lr}
  \footnotesize
  \setlength{\tabcolsep}{4pt}
  \begin{tabular}{@{}lccc@{}}
    \toprule
    Parameters & Initial LR & Final LR & Warm-up \\
    \midrule
    \makecell[l]{Skeleton offsets $\Delta\mathbf{z}$,\\and translational pose} & $10^{-3}$ & $10^{-4}$ & 1500 \\
    Pose offsets $\Delta\boldsymbol{\theta}_t$ & $10^{-4}$ & $2\times10^{-5}$ & 1500 \\
    \bottomrule
  \end{tabular}
\end{table}

\section{Datasets and evaluation}
\label{sec:supmat_dataeval}

\begin{table*}[t]
  \centering
  \caption{Evaluation settings. Cameras: F front, FL front left, FR front right, L and R the left and right KITTI cameras. Frame ranges include both the first and the last frame.}
  \label{tab:supmat_settings}
  \footnotesize
  \setlength{\tabcolsep}{4pt}
  \begin{tabular}{@{}l l c l l l l@{}}
    \toprule
    Setting & Data & Scenes & Cameras & Resolution & Held out for NVS & Results \\
    \midrule
    OmniRe split~\cite{chen2025omnire} & WOD & 8 & F, FL, FR & 960$\times$640 & every 10th frame & \cref{tab:omnire,tab:lidar,tab:ablations} \\
    \quad front camera & WOD & 8 & F & 960$\times$640 & every 10th frame & \cref{tab:omnire} \\
    \quad full resolution & WOD & 8 & F, FL, FR & 1920$\times$1280 & every 10th frame & \cref{tab:omnire} \\
    \quad sparse & WOD & 8 & F, FL, FR & 960$\times$640 & 50, 67 or 75\% of frames (evenly spaced) & \cref{tab:sparse,tab:downstream} \\
    PandaSet~\cite{lindstrom2026idsplat} & PandaSet & 10 & all six & 1920$\times$1080 & every 2nd frame & \cref{tab:pandaset} \\
    KITTI PVG~\cite{chen2026periodic} & KITTI-MOT & 3 & L, R & 1242$\times$375 & every 4th frame & \cref{tab:kitti} \\
    KITTI UnIRe~\cite{mao2026unire} & KITTI Raw & 4 & L, R & 621$\times$187 & every 10th frame & \cref{tab:kitti} \\
    CoDa-4DGS~\cite{song2025coda} & WOD, dynamic NOTR & 32 & F, FL, FR & 960$\times$640 & -- (reconstruction) & \cref{tab:coda} \\
    SplatFlow~\cite{sun2025splatflow} & WOD, PVG split & 4 & F, FL, FR & 960$\times$640 & every 4th frame & \cref{tab:sf} \\
    \bottomrule
  \end{tabular}
\end{table*}

\Cref{tab:supmat_settings} summarizes the settings. Reconstruction is evaluated on all training frames and NVS on the held-out frames, with the cameras and lidar sweeps of those frames held out. At held-out frames, instance trajectories are interpolated between the neighboring training frames, and pedestrian poses as described in \cref{sec:supmat_impl}.

\parsection{OmniRe setting (WOD)}
The eight Waymo~\cite{sun2020waymo} scenes of OmniRe~\cite{chen2025omnire} and UnIRe~\cite{mao2026unire} are listed in \cref{tab:supmat_omnire_scenes}. NVS holds out the frames at positions 10, 20, 30, \ldots{} of each range, counted from its first frame, for all three cameras and the lidar. For the sparse settings (\cref{sec:supmat_interp}), every second frame (50\%), every third frame (33\%), or every fourth frame (25\%) are used for training and all other frames are held out.

\begin{table}[t]
  \centering
  \caption{The eight Waymo scenes of the OmniRe split. Frame ranges include both the first and the last frame.}
  \label{tab:supmat_omnire_scenes}
  \footnotesize
  \setlength{\tabcolsep}{3pt}
  \begin{tabular}{@{}r l c@{}}
    \toprule
    Scene & Segment & Frames \\
    \midrule
    23  & 10455472356147194054\_1560\_000\_1580\_000 & 0--149 \\
    114 & 12505030131868863688\_1740\_000\_1760\_000 & 0--149 \\
    172 & 1382515516588059826\_780\_000\_800\_000 & 30--179 \\
    327 & 16951470340360921766\_2840\_000\_2860\_000 & 0--149 \\
    552 & 4487677815262010875\_4940\_000\_4960\_000 & 0--139 \\
    621 & 5846229052615948000\_2120\_000\_2140\_000 & 0--139 \\
    703 & 7761658966964621355\_1000\_000\_1020\_000 & 0--169 \\
    788 & 9653249092275997647\_980\_000\_1000\_000 & 130--198 \\
    \bottomrule
  \end{tabular}
\end{table}

\parsection{Single camera (WOD)}
AD-GS~\cite{xu2025ad} supports only the front camera, so we compare with it on the OmniRe split with the front camera alone, with the same frames and held-out frames.

\parsection{Full resolution (WOD)}
We also evaluate our method at full resolution (1920$\times$1280) to support future comparisons at higher fidelity.

\parsection{PandaSet}
We follow the protocol of IDSplat and use the PandaSet~\cite{xiao2021pandaset} sequences 001, 011, 016, 028, 053, 063, 084, 106, 123 and 158, with all six cameras at full resolution (1920$\times$1080) and every second frame held out.

\parsection{KITTI}
We evaluate on KITTI with two protocols. The protocol of PVG~\cite{chen2026periodic} uses the KITTI-MOT~\cite{geiger2012kitti} sequences 0001 (frames 380--431), 0002 (140--224) and 0006 (65--120) with both cameras at full resolution and holds out every fourth frame, starting from the fourth (positions 3, 7, 11, \ldots{} of each range, PVG's \emph{testhold} of 4). The protocol of UnIRe~\cite{mao2026unire} uses the KITTI Raw~\cite{geiger2012kitti} drives 0001 (all 108 frames), 0018, 0059 and 0091 (frames 0--199) of 2011\_09\_26, taken from UnIRe's released configuration, with both cameras at half resolution (621$\times$187), and holds out the frames at positions 10, 20, 30, \ldots. Moving pedestrians appear only in drives 0059 and 0091. 

\parsection{CoDa setting (dynamic NOTR)}
As in the released code of CoDa-4DGS~\cite{song2025coda}, we use frames 0--49 of the 32 dynamic scenes of NOTR~\cite{yang2023emernerf}, with the three front cameras at 960$\times$640, and evaluate reconstruction only.

\parsection{SplatFlow setting (WOD)}
SplatFlow~\cite{sun2025splatflow} uses four of the Waymo scenes of PVG~\cite{chen2026periodic}: 17 (frames 61--108), 22 (26--73), 50 (6--53) and 81 (26--73). We use them with the three front cameras at 960$\times$640 and hold out every fourth frame. Pedestrians move in scenes 50 and 81.

\parsection{Metrics}
PSNR and SSIM~\cite{wang2004ssim} are computed on the full image, and LPIPS~\cite{zhang2018lpips} with AlexNet~\citeS{krizhevsky2012alexnet} features, except on KITTI and in the CoDa and SplatFlow settings, where our rows and those of IDSplat use VGG features, as the evaluation code of PVG~\cite{chen2026periodic}. 
DPSNR is the PSNR computed only within the image regions of moving objects, as in EmerNeRF~\cite{yang2023emernerf}. We compute it separately for each class. On Waymo, the region of a ground-truth 3D box is the 2D rectangle that bounds its eight projected corners, clipped to the image. Boxes with a horizontal speed of at most 1\,m/s, without lidar points (likely occluded), or with a corner that cannot be projected are skipped. On PandaSet and KITTI, we compute the same regions from the boxes of each dataset, at the image time, and skip a box if a corner lies behind the camera or no lidar point of the nearest sweep falls inside it. For each class, DPSNR is averaged over the images in which it occurs and then over the scenes. The CoDa setting only reports DPSNR over all moving objects. The boxes are used only for evaluation.

The lidar metrics follow SplatAD~\cite{hess2025splatad}. The metrics consist of the median squared depth error and the RMSE of the intensity over returned rays, the accuracy of the predicted ray drop over all rays, and the Chamfer distance, the sum of the squared distances from each point to the nearest point of the other point cloud in both directions, divided by the number of measured points. All are computed on the held-out sweeps and averaged over the sweeps and then the scenes.  

\emph{Pedestrian points:} 
For the pedestrian columns of \cref{tab:lidar}, we evaluate the lidar rays that hit pedestrians, which are defined as the rays whose measured point lies inside a ground-truth pedestrian box of the sweep, enlarged by 0.1\, m. We evaluate only the pedestrians that our method reconstructs. At each held-out frame, every track present at that time is matched one-to-one to the nearest bounding box whose center is within 1\,m of it. The Chamfer distance is computed separately for each box and averaged over the boxes with rendered points. The metric is thus averaged per pedestrian rather than per sweep. 

\parsection{Baselines}
We provide more detail about the baselines used in this work. 

\emph{OmniRe Setting:}
The rows of OmniRe, UnIRe with and without ground-truth boxes, EmerNeRF, PVG and DeSiRe-GS are from UnIRe~\cite{mao2026unire}. SplatAD~\cite{hess2025splatad} is trained with ground-truth boxes in its published configuration, in which pedestrians are not modeled as actors. DIAL-GS~\cite{su2025dial} is trained with the authors' Waymo configuration, and IDSplat~\cite{lindstrom2026idsplat} with its released code and Grounded-SAM~2~\cite{ren2024groundedsam2} masks. AD-GS~\cite{xu2025ad} is trained with its released code on the front camera. These runs are scored by us with the same images, regions and metrics as our method. For the qualitative comparisons (\cref{sec:supmat_qualitative}), the instance decomposition and the downstream evaluation, which need rendered images, we run UnIRe with its released code, without ground-truth boxes; over the eight scenes, our runs reach an NVS PSNR of 30.29\,dB, against 31.56\,dB in its paper, with scene 114 the furthest below (26.65\,dB). 

\emph{PandaSet:}
The baseline rows are from IDSplat. We reran the IDSplat experiments with its released code and only added the pedestrian DPSNR calculation. 

\emph{KITTI:}
The rows of the PVG protocol are from DIAL-GS~\cite{su2025dial}, which repeats those of DeSiRe-GS~\cite{peng2025desire} and PVG~\cite{chen2026periodic}. AD-GS results are from its paper. The rows of the UnIRe protocol are from UnIRe. The released IDSplat code has no KITTI loader, so the IDSplat rows are using the IDSplat model but the KITTI dataloading and preprocessing is our implementation. We use SAM~3 masks for both IDSplat and our method in this table.   

\emph{CoDa and SplatFlow Settings:}
The baseline rows in \cref{tab:coda} and \cref{tab:sf} are from CoDa-4DGS~\cite{song2025coda} and from SplatFlow~\cite{sun2025splatflow}, respectively; IDSplat is run by us with its released code and Grounded-SAM~2 masks.

\section{Additional results}
\label{sec:supmat_results}
\Cref{tab:coda,tab:sf,tab:kitti} give the results in the further settings of \cref{sec:supmat_dataeval}. In these tables, only metrics reported by the baselines are given. Since the background and the vehicles keep the representation of IDSplat, the posable bodies can change the results only where pedestrians move. The OmniRe scenes were selected since they include many pedestrians. The settings in this section include fewer, and thus we expect full-image metrics closer to those of IDSplat and smaller gains on pedestrian-specific metrics than in \cref{tab:omnire}.

\parsection{CoDa setting (Dynamic NOTR)}
\begin{table}[t]
  \centering
  \caption{Reconstruction in the CoDa setting~\cite{song2025coda} (dynamic NOTR, three front cameras, frames 0--49). DPSNR over all moving objects; PSNR and DPSNR in dB. We use VGG for LPIPS here. All methods are annotation-free. \colorbox{tabfirst}{First}, \colorbox{tabsecond}{second}, \colorbox{tabthird}{third}.}
  \label{tab:coda}
  \footnotesize
  \setlength{\tabcolsep}{3pt}
  \begin{tabular}{@{}l cccc@{}}
    \toprule
    Method & PSNR$\uparrow$ & SSIM$\uparrow$ & LPIPS$\downarrow$ & DPSNR$\uparrow$ \\
    \midrule
    EmerNeRF~\cite{yang2023emernerf}        & 28.16 & 0.806 & 0.228 & 24.32 \\
    4DGS~\citeS{wu20244dgs}                  & 31.04 & 0.908 & 0.124 & 26.88 \\
    S3Gaussian~\citeS{huang2024s3gaussian}   & 32.39 & 0.920 & \cellcolor{tabthird}0.097 & 28.39 \\
    CoDa-4DGS~\cite{song2025coda}           & \cellcolor{tabthird}32.86 & \cellcolor{tabthird}0.927 & \cellcolor{tabfirst}0.086 & \cellcolor{tabthird}29.25 \\
    IDSplat (rigid)~\cite{lindstrom2026idsplat} & \cellcolor{tabsecond}36.11 & \cellcolor{tabsecond}0.964 & 0.098 & \cellcolor{tabsecond}34.53 \\
    \modelname{} (ours)                                   & \cellcolor{tabfirst}36.33 & \cellcolor{tabfirst}0.965 & \cellcolor{tabsecond}0.096 & \cellcolor{tabfirst}35.03 \\
    \bottomrule
  \end{tabular}
\end{table}
Our method and IDSplat reconstruct these scenes with higher image quality than the methods that model dynamic objects without instances, including CoDa-4DGS, and this difference is larger in the dynamic regions of the images. Our method is slightly ahead of IDSplat over all moving objects, likely due to the pedestrians in the scenes, though in NOTR the pedestrians are often a small part of the images. 

\parsection{SplatFlow setting (PVG scenes)}
\begin{table*}[t]
  \centering
  \caption{Reconstruction and NVS in the SplatFlow setting~\cite{sun2025splatflow} (the 4 Waymo scenes of PVG~\cite{chen2026periodic}, three front cameras, every 4th frame held out). PSNR in dB. We use VGG for LPIPS here. \colorbox{tabfirst}{First}, \colorbox{tabsecond}{second}, \colorbox{tabthird}{third}.}
  \label{tab:sf}
  \footnotesize
  \setlength{\tabcolsep}{4pt}
  \begin{tabular}{@{}l c c ccc @{\hskip 8pt} ccc@{}}
    \toprule
    & & & \multicolumn{3}{c}{Reconstruction} & \multicolumn{3}{c}{NVS} \\
    \cmidrule(lr){4-6} \cmidrule(l){7-9}
    Method & \makecell{Anno.\\free} & \makecell{Inst.\\decomp.} & PSNR$\uparrow$ & SSIM$\uparrow$ & LPIPS$\downarrow$ & PSNR$\uparrow$ & SSIM$\uparrow$ & LPIPS$\downarrow$ \\
    \midrule
    EmerNeRF~\cite{yang2023emernerf}        & \cmark & \xmark & 28.11 & 0.786 & 0.373 & 25.92 & 0.763 & 0.384 \\
    PVG~\cite{chen2026periodic}             & \cmark & \xmark & 32.46 & 0.910 & 0.229 & 28.11 & 0.849 & 0.279 \\
    DeSiRe-GS~\cite{peng2025desire}         & \cmark & \xmark & 33.61 & 0.919 & 0.204 & \cellcolor{tabthird}29.75 & \cellcolor{tabsecond}0.878 & \cellcolor{tabsecond}0.213 \\
    SplatFlow~\cite{sun2025splatflow}       & \cmark & \xmark & \cellcolor{tabthird}33.64 & \cellcolor{tabthird}0.951 & \cellcolor{tabthird}0.198 & 28.71 & \cellcolor{tabthird}0.874 & \cellcolor{tabthird}0.239 \\
    IDSplat~\cite{lindstrom2026idsplat} & \cmark & \cmark & \cellcolor{tabsecond}35.94 & \cellcolor{tabsecond}0.955 & \cellcolor{tabsecond}0.109 & \cellcolor{tabfirst}30.06 & \cellcolor{tabfirst}0.879 & \cellcolor{tabfirst}0.181 \\
    \modelname{} (ours)                                   & \cmark & \cmark & \cellcolor{tabfirst}35.99 & \cellcolor{tabfirst}0.956 & \cellcolor{tabfirst}0.107 & \cellcolor{tabsecond}30.04 & \cellcolor{tabsecond}0.878 & \cellcolor{tabfirst}0.181 \\
    \bottomrule
  \end{tabular}
\end{table*}
In \cref{tab:sf}, our full-image metrics match those of IDSplat, so adding posable pedestrians does not degrade the rest of the scene. As explained in \cref{sec:supmat_dataeval}, there are not many moving pedestrians in these scenes. 

\parsection{KITTI}
\begin{table*}[t]
  \centering
  \caption{Reconstruction and NVS on KITTI~\cite{geiger2012kitti}. PSNR in dB; our rows use LPIPS with VGG. AD-GS holds out every 4th frame from the 4th frame on and uses LPIPS with AlexNet. Colors rank the annotation-free methods. \colorbox{tabfirst}{First}, \colorbox{tabsecond}{second}, \colorbox{tabthird}{third}.}
  \label{tab:kitti}
  \footnotesize
  \setlength{\tabcolsep}{5pt}
  \begin{tabular}{@{}l c c ccc @{\hskip 6pt} ccc@{}}
    \toprule
    & & & \multicolumn{3}{c}{Reconstruction} & \multicolumn{3}{c}{NVS} \\
    \cmidrule(lr){4-6} \cmidrule(l){7-9}
    Method & \makecell{Anno.\\free} & \makecell{Inst.\\decomp.} & PSNR$\uparrow$ & SSIM$\uparrow$ & LPIPS$\downarrow$ & PSNR$\uparrow$ & SSIM$\uparrow$ & LPIPS$\downarrow$ \\
    \midrule
    \multicolumn{9}{@{}l}{\emph{KITTI-MOT, PVG protocol}} \\
    NSG~\citeS{ost2021neuralscenegraphs}     & \xmark & \cmark & 19.19 & 0.683 & 0.189 & 17.78 & 0.645 & 0.312 \\
    MARS~\citeS{wu2024mars}                 & \xmark & \cmark & 27.96 & 0.900 & 0.185 & 24.31 & 0.845 & 0.160 \\
    \dashrule{1-9}
    S-NeRF~\citeS{xie2023snerf}              & \cmark & \cmark & 19.23 & 0.664 & 0.193 & 18.71 & 0.606 & 0.352 \\
    StreetSurf~\citeS{guo2023streetsurf}     & \cmark & \xmark & 24.14 & 0.819 & 0.257 & 22.48 & 0.763 & 0.304 \\
    3DGS~\cite{kerbl20233dgs}               & \cmark & \xmark & 21.02 & 0.811 & 0.202 & 19.54 & 0.776 & 0.224 \\
    SUDS~\citeS{turki2023suds}               & \cmark & \xmark & 28.83 & 0.917 & 0.147 & 26.07 & 0.797 & 0.131 \\
    EmerNeRF~\cite{yang2023emernerf}        & \cmark & \xmark & 26.95 & 0.828 & 0.218 & 25.24 & 0.801 & 0.237 \\
    PVG~\cite{chen2026periodic}             & \cmark & \xmark & 32.83 & 0.937 & 0.070 & 27.43 & 0.896 & 0.114 \\
    DeSiRe-GS~\cite{peng2025desire}         & \cmark & \xmark & 33.94 & \cellcolor{tabthird}0.949 & \cellcolor{tabsecond}0.040 & \cellcolor{tabthird}28.87 & \cellcolor{tabthird}0.901 & 0.106 \\
    DIAL-GS~\cite{su2025dial}               & \cmark & \cmark & \cellcolor{tabthird}34.23 & \cellcolor{tabsecond}0.954 & \cellcolor{tabfirst}0.039 & 28.34 & \cellcolor{tabsecond}0.911 & \cellcolor{tabfirst}0.084 \\
    AD-GS~\cite{xu2025ad}         & \cmark & \xmark & --    & --    & --    & 29.16 & 0.920 & 0.033 \\
    IDSplat~\cite{lindstrom2026idsplat} & \cmark & \cmark & \cellcolor{tabfirst}34.35 & \cellcolor{tabfirst}0.959 & \cellcolor{tabthird}0.056 & \cellcolor{tabfirst}29.82 & \cellcolor{tabfirst}0.922 & \cellcolor{tabsecond}0.094 \\
    \modelname{} (ours)                                   & \cmark & \cmark & \cellcolor{tabsecond}34.31 & \cellcolor{tabfirst}0.959 & 0.057 & \cellcolor{tabsecond}29.79 & \cellcolor{tabfirst}0.922 & \cellcolor{tabthird}0.095 \\
    \midrule
    \multicolumn{9}{@{}l}{\emph{KITTI Raw, UnIRe protocol}} \\
    HUGS~\cite{zhou2024hugs}                & \xmark & \cmark & 27.14 & 0.908 & 0.082 & 23.91 & 0.750 & 0.094 \\
    StreetGS~\cite{yan2024streetgs}         & \xmark & \cmark & 27.59 & 0.910 & 0.065 & 24.15 & 0.793 & 0.084 \\
    OmniRe~\cite{chen2025omnire}            & \xmark & \cmark & 28.22 & 0.916 & 0.072 & 26.52 & 0.881 & 0.081 \\
    \dashrule{1-9}
    EmerNeRF~\cite{yang2023emernerf}        & \cmark & \xmark & 26.95 & 0.831 & 0.197 & 25.11 & 0.801 & 0.227 \\
    PVG~\cite{chen2026periodic}             & \cmark & \xmark & 27.40 & 0.895 & 0.097 & 24.34 & 0.819 & 0.121 \\
    DeSiRe-GS~\cite{peng2025desire}         & \cmark & \xmark & 28.62 & \cellcolor{tabthird}0.921 & 0.085 & 25.32 & \cellcolor{tabthird}0.846 & \cellcolor{tabthird}0.096 \\
    UnIRe~\cite{mao2026unire}               & \cmark & \cmark & \cellcolor{tabthird}28.92 & \cellcolor{tabsecond}0.929 & \cellcolor{tabthird}0.064 & \cellcolor{tabthird}26.10 & \cellcolor{tabsecond}0.884 & \cellcolor{tabsecond}0.079 \\
    IDSplat~\cite{lindstrom2026idsplat} & \cmark & \cmark & \cellcolor{tabsecond}33.98 & \cellcolor{tabfirst}0.971 & \cellcolor{tabfirst}0.043 & \cellcolor{tabsecond}29.04 & \cellcolor{tabfirst}0.924 & \cellcolor{tabfirst}0.078 \\
    \modelname{} (ours)                                   & \cmark & \cmark & \cellcolor{tabfirst}34.08 & \cellcolor{tabfirst}0.971 & \cellcolor{tabsecond}0.044 & \cellcolor{tabfirst}29.14 & \cellcolor{tabfirst}0.924 & \cellcolor{tabfirst}0.078 \\
    \bottomrule
  \end{tabular}
\end{table*}
Moving pedestrians are rare on KITTI. The three KITTI-MOT sequences contain a single standing pedestrian, and of the four KITTI Raw drives only 0059 and 0091 contain moving pedestrians. The comparison (\cref{tab:kitti}) therefore mainly tests whether the extended representation affects the reconstruction of the rest of the scene. In both protocols, our results are on par with IDSplat, and both are ahead of the other annotation-free methods in PSNR and SSIM.

\section{Qualitative results}
\label{sec:supmat_qualitative}
\begin{figure*}[t]
  \centering
  \includegraphics[width=0.88\textwidth]{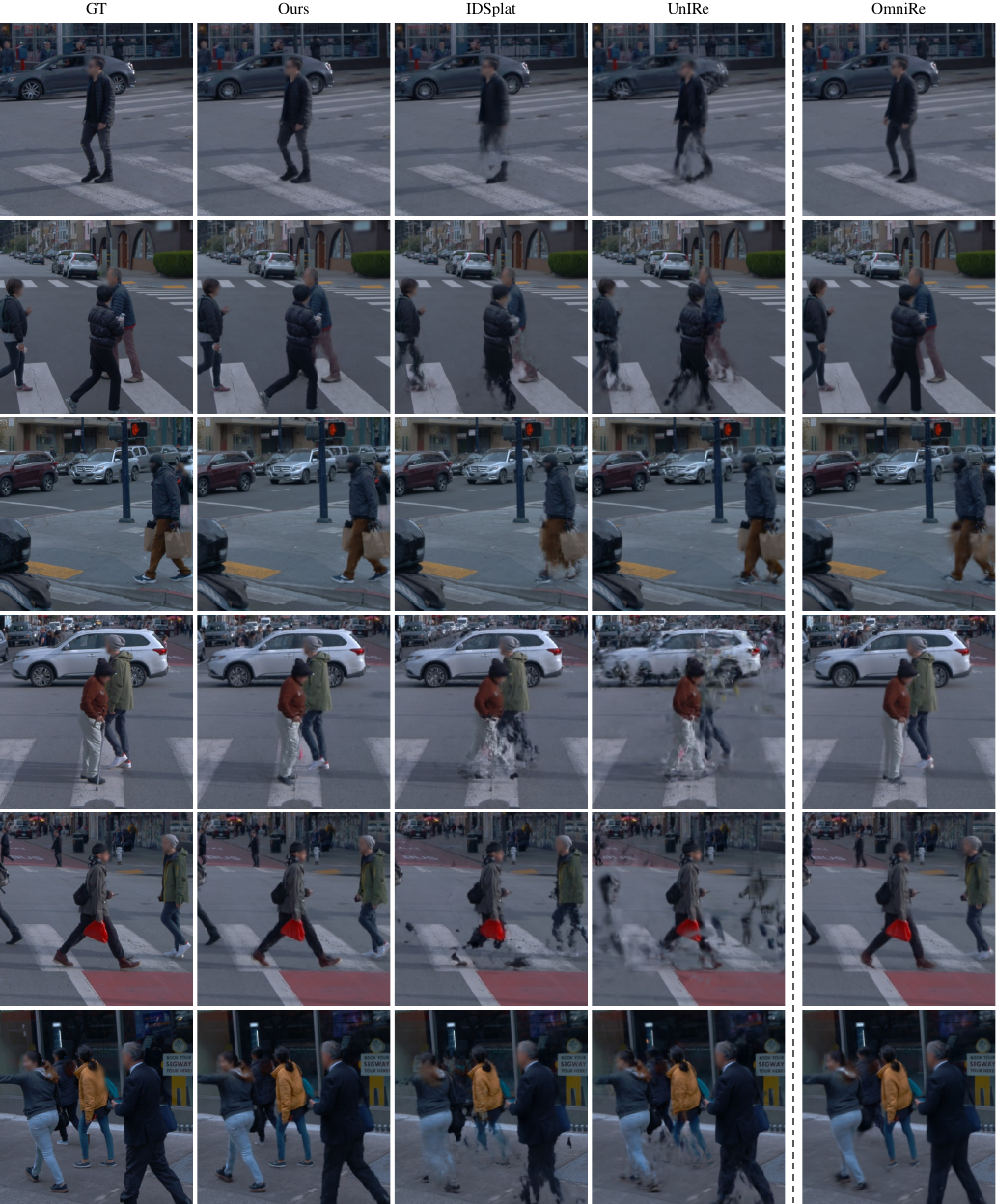}
  \caption{NVS of pedestrians at held-out frames of the WOD OmniRe split, as in \cref{fig:quality}. OmniRe uses ground-truth boxes.}
  \label{fig:supmat_quality_a}
\end{figure*}

\begin{figure*}[t]
  \centering
  \includegraphics[width=0.48\textwidth]{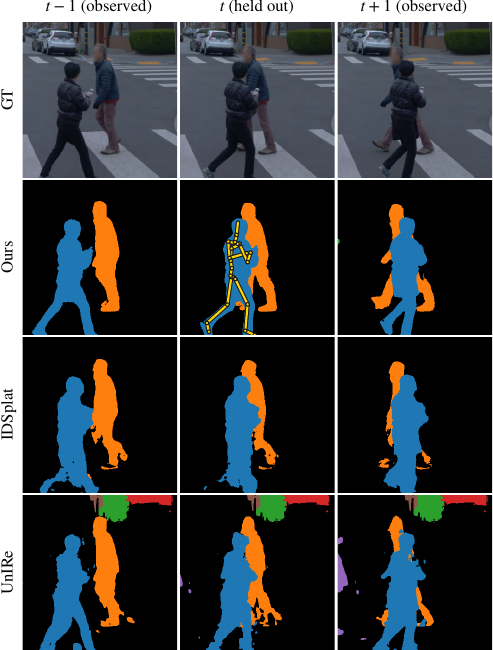}\hfill
  \includegraphics[width=0.48\textwidth]{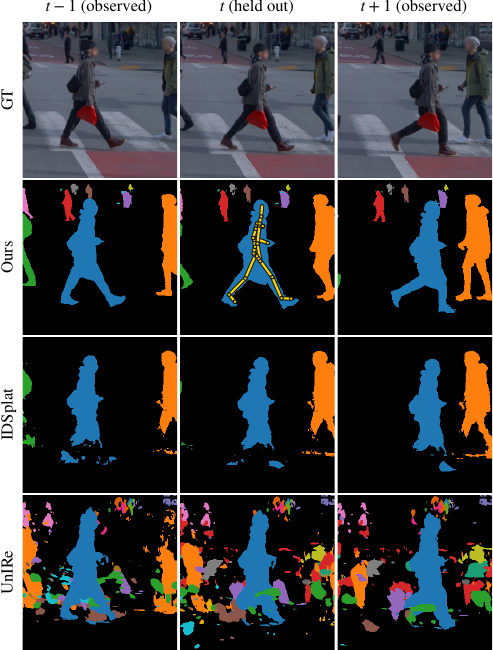}
  \caption{Rendered instance masks at a held-out frame $t$ and at the neighboring training frames, with our projected \mhr{} skeleton at $t$, as in \cref{fig:decomposition}. Left: scene 23, front camera. Right: scene 788, front camera.}
  \label{fig:supmat_masks}
\end{figure*}

\Cref{fig:supmat_quality_a} shows further qualitative results on WOD. We also present more examples of instance decomposition in \cref{fig:supmat_masks} and in a video uploaded together with this supplementary material (\texttt{instance\_decomposition\_all\_scenes.mp4}). The video shows the reconstructed images and instance decomposed 3D Gaussians of all 8 scenes in the OmniRe subset.  

\section{Learned interpolation}
\label{sec:supmat_interp}

We investigate how the reconstructed pedestrians behave when fewer training frames are available. Although the shared skeleton keeps the body geometry consistent across time, the poses between observations must be inferred, and the gaps between observations grow as frames are removed. We therefore train on 50\%, 33\% and 25\% of the frames of the OmniRe split (\cref{sec:supmat_dataeval}) and render the same models at the held-out frames with interpolation alone, with the correction network (\cref{sec:supmat_impl}) and with CondMDI~\cite{cohan2024condmdi} in its place (\cref{tab:sparse}). Besides the rendering metrics, we measure the distance between the joints of the rendered pose and those of the pose that \sambody{} estimates independently at the held-out frame, relative to the pelvis. 

With half of the frames, most gaps last one frame, linear interpolation already approximates the intermediate motion well, and the correction network changes the rendering metrics little. As the gaps grow, interpolation increasingly misses the motion between observations, most visibly as legs that stay together within a step, and the correction network recovers part of it: with a third and a quarter of the frames, it improves pedestrian DPSNR and lowers the joint errors, most for the feet. The joint errors are measured against independent \sambody{} estimates, not against ground-truth poses, so they show how close the poses are to those of the estimator rather than to the true motion. That the rendering metrics, which compare with the images, improve consistently in the same settings indicates that the correction moves the poses towards the observed motion. The comparison therefore favors the correction network, which is trained to reproduce \sambody{} estimates.

CondMDI is a pretrained diffusion model for motion in-betweening, which we apply zero-shot. For each gap, a 2\,s window of the observed poses around it is converted to the HumanML3D representation of CondMDI at 20\,fps, with all observations as keyframes, and one sample is generated. Its correction at the held-out time is its own pose there minus the linear interpolation of its output between the two surrounding observations, computed for the 22 joints matched by name to those of SMPL; every other \mhr{} joint takes the correction of the nearest matched joint, and it is applied to the joint positions for the same gaps and pedestrians as our correction network.
At every density, CondMDI changes the results much less than our correction network. A likely reason is that a model trained on clean motion capture stays close to the given keyframes and does not complete the steps that noisy 10\,Hz observations miss. The matching between the \mhr{} and SMPL skeletons may also introduce errors.

\begin{figure}[t]
  \centering
  \includegraphics[width=\linewidth]{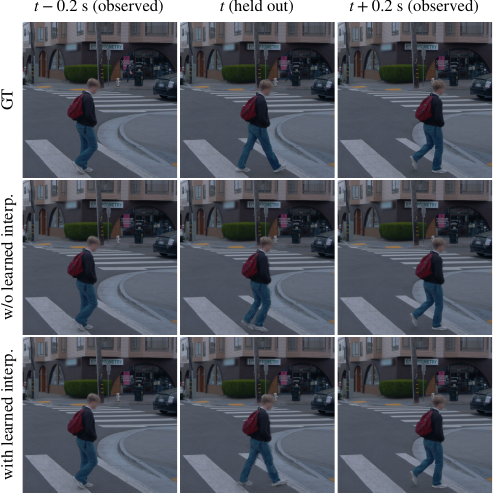}
  \caption{Scene 23, front camera, 25\% of training frames.}
  \label{fig:supmat_interp_023}
\end{figure}

\begin{figure}[t]
  \centering
  \includegraphics[width=\linewidth]{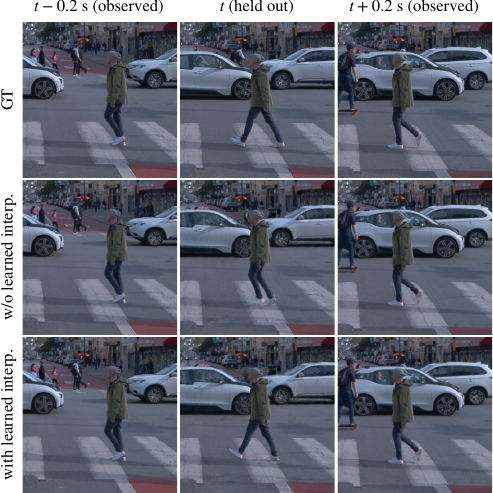}
  \caption{Scene 788, front camera, 25\% of training frames.}
  \label{fig:supmat_interp_788}
\end{figure}

\begin{figure}[t]
  \centering
  \includegraphics[width=\linewidth]{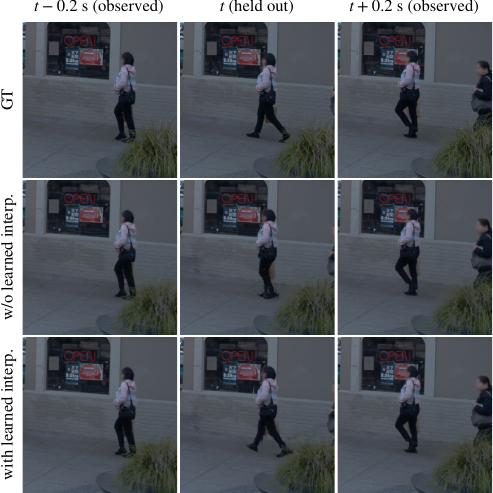}
  \caption{Scene 23, front right camera, 25\% of training frames.}
  \label{fig:supmat_interp_023b}
\end{figure}

\begin{figure}[t]
  \centering
  \includegraphics[width=\linewidth]{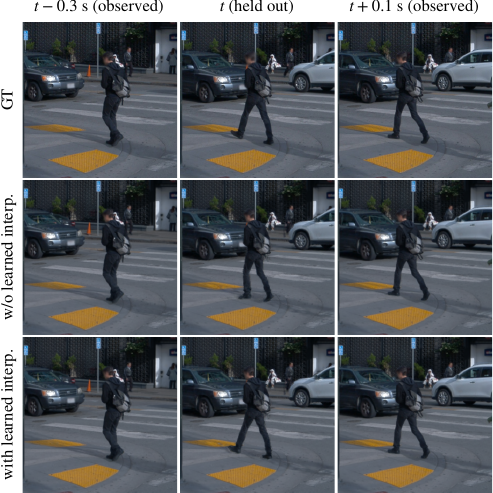}
  \caption{Scene 788, front left camera, 33\% of training frames.}
  \label{fig:supmat_interp_788b}
\end{figure}

In \cref{fig:supmat_interp_023,fig:supmat_interp_023b,fig:supmat_interp_788,fig:supmat_interp_788b}, we provide more qualitative examples of the effect of our learned interpolation approach. 

\begin{table}[t]
  \centering
  \caption{Novel view synthesis on the OmniRe split with fewer training frames.
    Each group of rows renders the same models, with poses between observations interpolated, corrected by our correction network, or corrected by CondMDI~\cite{cohan2024condmdi}.
    PSNR and DPSNR in dB.
    Joint error is measured on walking pedestrians at held-out frames, against the body estimated in that frame, for the feet, the knees, the hands and the mean over all 127 joints.}
  \label{tab:sparse}
  \footnotesize
  \setlength{\tabcolsep}{1.0pt}
  \begin{tabular}{@{}c l cc cccc@{}}
    \toprule
    & & & & \multicolumn{4}{c}{Joint error (cm)$\downarrow$} \\
    \cmidrule(l){5-8}
    Frames & Pose origin & PSNR$\uparrow$ & \makecell{DPSNR$\uparrow$\\ped.} & Feet & Knees & Hands & All \\
    \midrule
    \multirow{3}{*}{50\%} & Interpolation & 33.69 & 27.35 & 10.57 & 7.36 & 9.73 & 8.59 \\
                          & + correction network & 33.68 & 27.37 & 10.00 & 7.06 & 9.64 & 8.46 \\
                          & + CondMDI & 33.71 & 27.38 & 10.52 & 7.33 & 9.76 & 8.61 \\
    \midrule
    \multirow{3}{*}{33\%} & Interpolation & 32.29 & 25.54 & 12.16 & 8.16 & 10.28 & 9.17 \\
                          & + correction network & 32.48 & 25.87 & 10.54 & 7.36 & 10.06 & 8.80 \\
                          & + CondMDI & 32.32 & 25.58 & 12.16 & 8.16 & 10.31 & 9.19 \\
    \midrule
    \multirow{3}{*}{25\%} & Interpolation & 31.30 & 24.52 & 13.93 & 9.04 & 11.06 & 9.89 \\
                          & + correction network & 31.61 & 25.03 & 11.12 & 7.68 & 10.73 & 9.27 \\
                          & + CondMDI & 31.31 & 24.53 & 14.03 & 9.09 & 11.13 & 9.94 \\
    \bottomrule
  \end{tabular}
\end{table}

\section{Downstream perception}
\label{sec:supmat_downstream}
The downstream evaluation (\cref{tab:downstream}) uses every held-out frame of the three cameras in each sparse split of the OmniRe scenes (\cref{sec:supmat_dataeval}): 1{,}656 images with 22{,}931 Waymo 2D pedestrian boxes at 50\% of the frames, 2{,}235 images with 30{,}939 boxes at 33\%, and 2{,}508 images with 34{,}695 boxes at 25\%. Our renderings use the correction network (\cref{sec:supmat_impl}), and the row without the learned interpolation renders the same models with interpolation alone. IDSplat uses its released code and Grounded-SAM~2 masks, and UnIRe its released code without ground-truth boxes, both trained by us on the same frames. The real images are resized to the render resolution of 960$\times$640 before inference.

\parsection{Detection}
ViTDet~\cite{li2022vitdet} is the Cascade Mask R-CNN with a ViT-H backbone trained on COCO~\cite{lin2014coco}, of which we keep the person class, and Grounding DINO~\cite{liu2024groundingdino} is the base model with the text prompt ``pedestrian.''. AP50 is the all-point average precision against the Waymo 2D pedestrian boxes at an IoU of 0.5, over the detections with a score of at least 0.05. DA follows ~\cite{lindstrom2024real2sim}. The detections with a score of at least 0.5 in the rendered and in the real image are matched greedily at an IoU of 0.5, and each match is weighted by the ratio of the smaller to the larger box area. The weighted recall is computed in both directions, once with the real and once with the rendered detections as reference, and DA is the mean of the two, averaged over frames. A frame without detections in either image counts as full agreement. 

\parsection{Pose estimation}
ViTPose~\cite{xu2022vitpose} is ViTPose+ huge with its COCO head, applied to crops of the labeled Waymo 2D boxes of the pedestrians with keypoint labels for 3{,}732, 5{,}152 and 5{,}556 pedestrians at 50, 33 and 25\% of the frames. The 13 COCO keypoints with a Waymo counterpart are scored where Waymo marks them as not occluded. PCK~\cite{yang2013articulated} counts a keypoint as correct within 0.1 of the box height, and OKS~\cite{lin2014coco} uses the COCO keypoint constants with the box area as the object scale.

\section{Re-posing}
\label{sec:supmat_reposing}
To re-pose a pedestrian, we use the pose of another pedestrian in the scene. A target pedestrian takes the 124 joint angles of a source pedestrian frame by frame, and keeps its global translation and rotation, its translational parameters, its skeleton and its Gaussians, so the re-posed body has the bone lengths, position and heading of the target. We include more examples of reposing in two videos included in the supplementary material (\texttt{reposing\_scene621\_62\_takes\_71\_pose.mp4} and \texttt{reposing\_scene788\_42\_takes\_12\_pose.mp4}). 

\section{Runtime analysis}
\label{sec:supmat_runtime}
We measure the runtime on the OmniRe split in the NVS setting (\cref{tab:supmat_settings}), with one NVIDIA A100 GPU per scene, and report the median over the eight scenes (\cref{tab:supmat_runtime}). Preprocessing covers everything before the scene optimization: for our method and IDSplat the instance masks, the body estimates and the initialization of instances and Gaussians, and for the other methods their own inputs, such as optical flow, depth and semantic maps, camera tracking or, for DIAL-GS, its two first-stage optimizations. Evaluation during training and the conversion of the raw Waymo data are excluded. The inference rate is measured by rendering the held-out views after a warm-up and timing only the render call. For our method and IDSplat, only the RGB image is rendered. 

Optimization takes most of the time of our method (\cref{tab:supmat_runtime_components}). Most of the preprocessing is spent in the two pretrained models: SAM~3 runs one detection and tracking pass per text prompt and camera, and \sambody{} one estimate per pedestrian masklet and frame. The steps we add on top of them, masklet refinement, body alignment and skinning setup, take about two minutes per scene, and the steps inherited from IDSplat, from data loading to the initialization of the Gaussians, about a tenth of the total time. Compared with the other methods (\cref{tab:supmat_runtime}), our method takes longer than IDSplat, partly because the bodies are posed and deformed at every optimization step, about as long as AD-GS, longer than UnIRe and considerably less than DIAL-GS. At inference, posing the bodies, interpolating the \mhr{} parameters, applying the correction network and deforming the Gaussians, takes about half of the render time, which makes our method the slowest of the compared methods at about 19 frames per second at 960$\times$640.

\begin{table}[t]
  \centering
  \caption{Runtime on the OmniRe split (NVS) on one NVIDIA A100 GPU, median over the eight scenes. Preprocessing and optimization in minutes per scene; inference rate in megapixels per second. $^\dagger$AD-GS supports only the front camera and is run on it alone; the other methods use the three front cameras.}
  \label{tab:supmat_runtime}
  \footnotesize
  \setlength{\tabcolsep}{4pt}
  \begin{tabular}{@{}l r r r r@{}}
    \toprule
    Method & Preproc. & Optim. & Total & MP/s \\
    \midrule
    UnIRe~\cite{mao2026unire} & 44.1 & 47.2 & 91.2 & 14.5 \\
    AD-GS$^\dagger$~\cite{xu2025ad} & 40.2 & 85.6 & 128.1 & 29.9 \\
    DIAL-GS~\cite{su2025dial} & 212.8 & 131.5 & 344.5 & 36.3 \\
    IDSplat~\cite{lindstrom2026idsplat} & 17.1 & 59.0 & 75.4 & 27.7 \\
    \modelname{} (ours) & 46.7 & 75.8 & 122.9 & 11.7 \\
    \bottomrule
  \end{tabular}
\end{table}

\begin{table}[t]
  \centering
  \caption{Runtime of our method per component: minutes per scene (median over the eight scenes of the OmniRe split, so the components do not add up exactly to the total) and share of the total time.}
  \label{tab:supmat_runtime_components}
  \footnotesize
  \setlength{\tabcolsep}{4pt}
  \begin{tabular}{@{}l r r@{}}
    \toprule
    Component & Time (min) & Share (\%) \\
    \midrule
    Instance masks (SAM~3) & 12.3 & 10.0 \\
    Masklet refinement & 0.9 & 0.7 \\
    \sambody{} & 19.6 & 15.9 \\
    Data loading & 5.0 & 4.0 \\
    Point painting and instance merging & 1.0 & 0.8 \\
    Body alignment & 0.8 & 0.6 \\
    Registration & 4.4 & 3.6 \\
    Smoothing & 1.4 & 1.2 \\
    Gaussian initialization & 1.4 & 1.1 \\
    Skinning setup & 0.2 & 0.2 \\
    Optimization & 75.8 & 61.6 \\
    \midrule
    Total & 122.9 & 100.0 \\
    \bottomrule
  \end{tabular}
\end{table}

\section{Limitations}
\label{sec:supmat_limitations}
We discuss the limitations of the main paper in more detail. 

\begin{figure}[t]
  \centering
  \includegraphics[width=\linewidth]{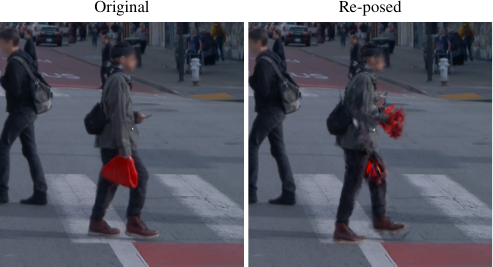}
  \caption{Re-posing a pedestrian who carries a bag (scene 788, front camera). The bag is not part of the body model, so its Gaussians follow the nearest body parts: when the arm and leg take the new pose, the bag splits between the hand and the leg.}
  \label{fig:supmat_repose_limits}
\end{figure}

\parsection{Carried objects and loose clothing} 
We do not explicitly model clothing or accessories and the body model describes the body only. Gaussians of bags, carried objects and loose clothing, which are initialized from lidar points, inherit the skinning weights of their nearest body part. At observed poses, they are optimized to match the images, but under a new pose they follow that body part, which is sometimes not how those objects are meant to move. \Cref{fig:supmat_repose_limits} shows the result. 

\begin{figure}[t]
  \centering
  \includegraphics[width=\linewidth]{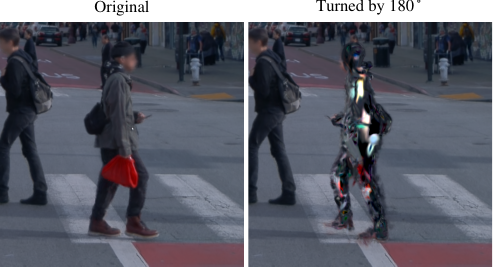}
  \caption{Turning a pedestrian by 180\textdegree{} about its vertical axis (scene 788, front camera). The camera now sees surfaces that were not observed during recording, and the view-dependent colors of the Gaussians, which do not turn with the pedestrian, are evaluated for directions they were not fitted to.}
  \label{fig:supmat_turn}
\end{figure}

\parsection{Unobserved surfaces} 
Gaussians are optimized only where a pedestrian was observed. Changing a pedestrian's orientation or pose may expose surfaces that were occluded or never seen. \Cref{fig:supmat_turn} shows a pedestrian turned to face the opposite direction. Edits are therefore most reliable when the pedestrian keeps the observed side towards the camera. 

\begin{figure}[t]
  \centering
  \includegraphics[width=\linewidth]{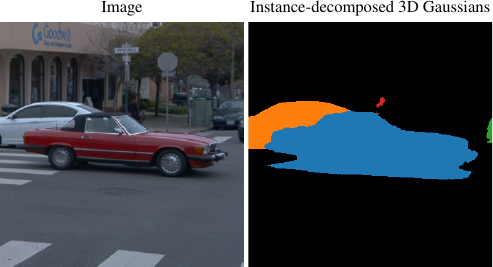}
  \caption{A car (scene 23, front-left camera, held-out frame) and the rendered instances (right). The instance of the car in front (blue) includes a flat region below it, where its shadow falls on the road. The shadow is represented by the car's Gaussians.}
  \label{fig:supmat_shadow}
\end{figure}

\parsection{Shadows and reflections} 
Shadows that a moving object casts, and its reflections in windows or wet roads, move with the object, so the optimization can explain them with the Gaussians of its instance, near its surface. Such effects are then moved with the object when it is edited, instead of following the scene lighting, and are left behind in the background where the object was removed. Modeling them explicitly would require a representation of the lighting, which the method does not have. \Cref{fig:supmat_shadow} shows a car whose instance includes its shadow. 

\begin{figure}[t]
  \centering
  \includegraphics[width=\linewidth]{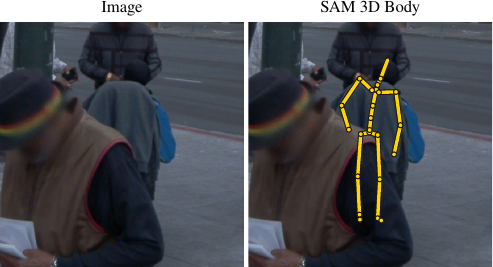}
  \caption{\sambody{} applied to a wheelchair with a jacket over it (scene 327, front-left camera). SAM~3 tracks it as a person, and \sambody{} fits a standing body to it (right, projected skeleton).}
  \label{fig:supmat_wheelchair}
\end{figure}

\parsection{Errors of the zero-shot models} 
The method relies on SAM~3 and \sambody{} without fine-tuning. A pedestrian that SAM~3 misses, or tracks in too few frames, is not reconstructed as an instance and remains part of the background, where it cannot be edited and is rendered blurred where it moves. The masklet refinement resolves duplicate and drifting masklets, but not missing ones. \sambody{} estimates each frame independently and sometimes confuses the front and the back of a pedestrian. The body refinement corrects some of these estimates, but where it does not, interpolating towards such an estimate produces a turn that the pedestrian did not make. These errors are more frequent for small, distant, or occluded pedestrians, which are observed in few pixels and lidar points and are also rendered less faithfully. For example, in \cref{fig:supmat_wheelchair}, a partially occluded person in a wheelchair is mistaken for a short standing pedestrian.

\section*{Supplementary references}
\begin{supprefs}
\item\label{suppref:kheradmand2024mcmc} Shakiba Kheradmand, Daniel Rebain, Gopal Sharma, Weiwei Sun, Yang-Che Tseng, Hossam Isack, Abhishek Kar, Andrea Tagliasacchi, and Kwang Moo Yi. 3D Gaussian splatting as Markov chain Monte Carlo. In \emph{Advances in Neural Information Processing Systems (NeurIPS)}, 2024.
\item\label{suppref:kingma2014adam} Diederik P. Kingma and Jimmy Ba. Adam: A method for stochastic optimization. In \emph{International Conference on Learning Representations (ICLR)}, 2015.
\item\label{suppref:loshchilov2019adamw} Ilya Loshchilov and Frank Hutter. Decoupled weight decay regularization. In \emph{International Conference on Learning Representations (ICLR)}, 2019.
\item\label{suppref:krizhevsky2012alexnet} Alex Krizhevsky, Ilya Sutskever, and Geoffrey E. Hinton. ImageNet classification with deep convolutional neural networks. In \emph{Advances in Neural Information Processing Systems (NeurIPS)}, volume 25, 2012. 
\item\label{suppref:wu20244dgs} Guanjun Wu, Taoran Yi, Jiemin Fang, Lingxi Xie, Xiaopeng Zhang, Wei Wei, Wenyu Liu, Qi Tian, and Xinggang Wang. 4D Gaussian splatting for real-time dynamic scene rendering. In \emph{IEEE/CVF Conference on Computer Vision and Pattern Recognition (CVPR)}, pages 20310--20320, 2024.
\item\label{suppref:huang2024s3gaussian} Nan Huang, Xiaobao Wei, Wenzhao Zheng, Pengju An, Ming Lu, Wei Zhan, Masayoshi Tomizuka, Kurt Keutzer, and Shanghang Zhang. $\textit{S}^3$Gaussian: Self-supervised street Gaussians for autonomous driving. In \emph{IEEE International Conference on Robotics and Automation (ICRA)}, 2026.
\item\label{suppref:ost2021neuralscenegraphs} Julian Ost, Fahim Mannan, Nils Thuerey, Julian Knodt, and Felix Heide. Neural scene graphs for dynamic scenes. In \emph{IEEE/CVF Conference on Computer Vision and Pattern Recognition (CVPR)}, pages 2856--2865, 2021.
\item\label{suppref:wu2024mars} Zirui Wu, Tianyu Liu, Liyi Luo, Zhide Zhong, Jianteng Chen, Hongmin Xiao, Chao Hou, Haozhe Lou, Yuantao Chen, Runyi Yang, Yuxin Huang, Xiaoyu Ye, Zike Yan, Yongliang Shi, Yiyi Liao, and Hao Zhao. MARS: An instance-aware, modular and realistic simulator for autonomous driving. In \emph{Artificial Intelligence}, pages 3--15. Springer Nature Singapore, 2024.
\item\label{suppref:xie2023snerf} Ziyang Xie, Junge Zhang, Wenye Li, Feihu Zhang, and Li Zhang. S-NeRF: Neural radiance fields for street views. In \emph{International Conference on Learning Representations (ICLR)}, 2023.
\item\label{suppref:guo2023streetsurf} Jianfei Guo, Nianchen Deng, Xinyang Li, Ye-Qi Bai, Botian Shi, Chi-Yu Wang, Chenjing Ding, Dongliang Wang, and Yi-Kang Li. StreetSurf: Extending multi-view implicit surface reconstruction to street views. \emph{arXiv preprint arXiv:2306.04988}, 2023.
\item\label{suppref:turki2023suds} Haithem Turki, Jason Y. Zhang, Francesco Ferroni, and Deva Ramanan. SUDS: Scalable urban dynamic scenes. In \emph{IEEE/CVF Conference on Computer Vision and Pattern Recognition (CVPR)}, 2023.
\end{supprefs}

\end{document}